\documentclass[sigconf]{acmart}

\usepackage{bm}

\def\secref#1{section~\ref{#1}}

\def\eqref#1{equation~\ref{#1}}

\def\1{\bm{1}}

\makeatletter
\@ifundefined{mathsfit}{%
  \DeclareMathAlphabet{\mathsfit}{\encodingdefault}{\sfdefault}{m}{sl}
  \SetMathAlphabet{\mathsfit}{bold}{\encodingdefault}{\sfdefault}{bx}{n}
}{}
\makeatother

\usepackage{adjustbox}
\usepackage{tcolorbox}
\usepackage{multirow}
\tcbuselibrary{breakable, skins}
\usepackage{fontawesome5}
\usepackage{algorithm}
\usepackage{algorithmic}
\usepackage{listings}
\usepackage{pdfpages}
\usepackage{afterpage}
\usepackage[capitalise]{cleveref}
\usepackage{enumitem}
\usepackage{siunitx}

\usepackage{tikz}
\usepackage{pgfplots}

\usetikzlibrary{arrows.meta}
\usetikzlibrary{calc}
\usetikzlibrary{positioning}
\usetikzlibrary{shapes.geometric}
\usetikzlibrary{decorations.pathmorphing}
\usetikzlibrary{intersections}
\usetikzlibrary{matrix}
\usetikzlibrary{backgrounds}
\usetikzlibrary{shapes.multipart,fit}

\pgfplotsset{compat=1.18}

\definecolor{darkblue}{rgb}{0, 0, 0.5}

\definecolor{exampleblue}{RGB}{52, 152, 219}
\definecolor{examplegray}{RGB}{245, 247, 250}
\definecolor{textgray}{RGB}{52, 58, 64}
\definecolor{bordercolor}{RGB}{168, 174, 179}

\newtcolorbox{example}[1][]{%
    enhanced,
    breakable,
    colback=examplegray,
    colframe=bordercolor,
    coltitle=white,
    coltext=textgray,
    boxrule=1pt,
    arc=4pt,
    outer arc=4pt,
    toptitle=3mm,
    bottomtitle=3mm,
    fonttitle=\sffamily\bfseries,
    title={\hspace{8pt}Contributions},
#1
}

\newtcolorbox{exampleblue}[1][]{%
  enhanced,
  breakable,
  colback=white,
  colframe=exampleblue,
  coltitle=white,
  coltext=textgray,
  boxrule=2pt,
  arc=4pt,
  outer arc=4pt,
  toptitle=3mm,
  bottomtitle=3mm,
  fonttitle=\sffamily\bfseries,
  title={\faRobot\hspace{8pt}Research Question},
  attach boxed title to top left={xshift=0pt,yshift=-\tcboxedtitleheight/2},
  boxed title style={%
      colback=exampleblue,
      arc=3pt,
      outer arc=3pt,
      boxrule=0pt
  },
  #1
}

\newtcolorbox{examplecompact}[1][]{%
  enhanced,
  breakable,
  colback=examplegray,
  colframe=bordercolor,
  coltext=textgray,
  boxrule=0.5pt,
  arc=3pt,
  outer arc=3pt,
  before upper={\small\faRobot\hspace{6pt}},
  #1
}

\copyrightyear{2026}
\acmYear{2026}
\setcopyright{cc}
\setcctype{by}
\acmConference[CAIS '26]{ACM Conference on AI and Agentic Systems}{May 26--29, 2026}{San Jose, CA, USA}
\acmBooktitle{ACM Conference on AI and Agentic Systems (CAIS '26), May 26--29, 2026, San Jose, CA, USA}
\acmDOI{10.1145/3786335.3813150}
\acmISBN{979-8-4007-2415-2/2026/05}



\author{Benjamin Feuer}
\affiliation{%
  \institution{Stanford University}
  \country{USA}
}
\email{bfeuer@stanford.edu}

\author{Lennart Purucker}
\affiliation{%
  \institution{Prior Labs}
  \country{Germany}
}
\email{puruckel@cs.uni-freiburg.de}

\author{Oussama Elachqar}
\affiliation{%
  \institution{Oumi}
  \country{USA}
}
\email{oussama@oumi.ai}

\author{Chinmay Hegde}
\affiliation{%
  \institution{New York University}
  \country{USA}
}
\email{chinmay.h@nyu.edu}

\title{MARVIS: Modality Adaptive Reasoning over VISualizations}

\begin{abstract}
Predictive applications of machine learning often rely on small (sub 1 Bn parameter) specialized models tuned to particular domains or modalities. Such models often achieve excellent performance, but lack flexibility. LLMs and VLMs offer versatility, but typically underperform specialized predictors, especially on non-traditional modalities and long-tail domains. We propose MARVIS (Modality Adaptive Reasoning over VISualizations), a system that transforms latent embedding spaces into visual representations and then leverages the spatial and fine-grained reasoning skills of VLMs to interpret the visualizations and utilize them for predictions successfully. MARVIS achieves competitive performance across vision, audio, biological, and tabular domains using a single 3B parameter model, yielding results that beat Gemini 2.0 by 16\% on average. MARVIS drastically reduces the gap between LLM/VLMs approaches and specialized domain-specific methods, without requiring any domain-specific training. Code and datasets are available at \url{https://github.com/penfever/marvis}.
\end{abstract}

\begin{document}

\maketitle

\begin{figure*}[!t]
\centering
\begin{minipage}{0.5\textwidth}
    \centering
    \includegraphics[width=\linewidth]{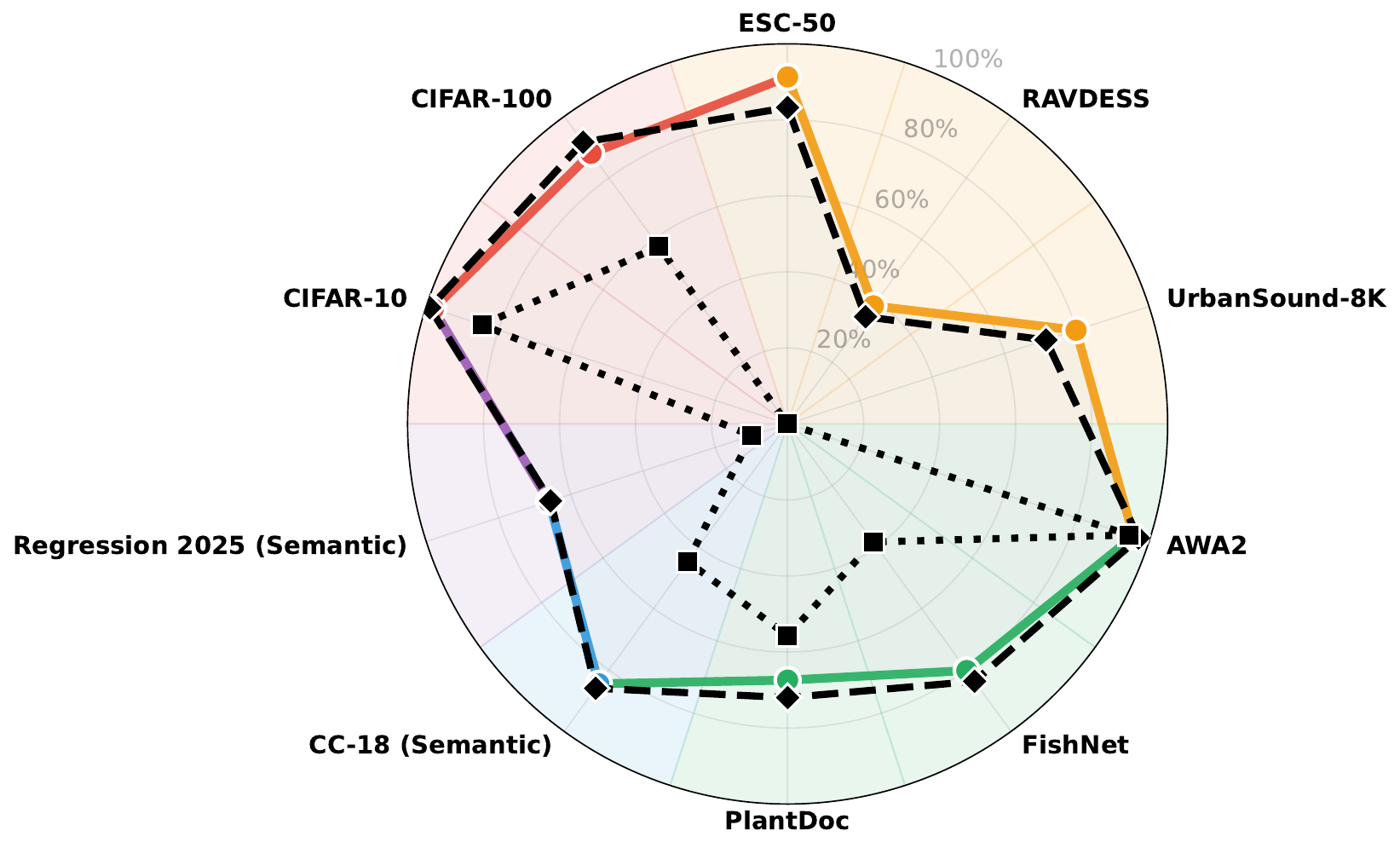}
\end{minipage}%
\begin{minipage}{0.18\textwidth}
    \centering
    \includegraphics[width=\linewidth]{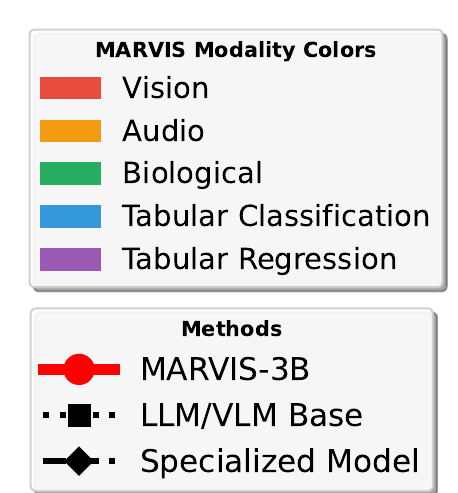}
\end{minipage}
\caption{\textbf{MARVIS transforms VLMs into frontier predictors.} Using a standard 3B parameter QwenVL model zero-shot without reasoning, MARVIS (colored line) achieves competitive performance compared to specialized baselines (dashed line) across modalities and domains, far exceeding the best existing LLM / VLM predictors (dotted line).
}
\label{fig:radar}
\end{figure*}

\section{Introduction}

Much of the progress in the field of machine learning in recent years has been on classification and regression tasks (which, in this work, we sometimes collectively refer to as \textit{predictive} tasks). These have historically been addressed either using classical machine learning methods or, more recently, with deep learning. In the latter case, the best performance has generally been achieved using \textbf{specialized models} with less than one billion parameters tuned for a particular task and/or knowledge domain~\citep{prokhorenkova2018catboost,he2015deepresiduallearningimage,hollmann_accurate_2025}. These models often learn to compress a high-dimensional input space into a simplified embedded space; in some cases, these embeddings can be used for prediction without any fine-tuned classification stage via classical nonparametric methods like KNN \citep{oquab2023dinov2}, but in most cases,  parametric fine-tuning follows. What specialist models gain in precision, however, they sacrifice in flexibility -- narrow experts are often inapplicable to other domains without additional fine-tuning~\citep{devlin2019bertpretrainingdeepbidirectional}.

\textbf{LLM and VLMs} introduced an exciting new paradigm: in-context learning (ICL) over text and images. Unlike specialist models, these so-called foundation models adapt to new tasks without weight updates~\citep{brown2020languagemodelsfewshotlearners}. Unlike specialists, LLMs are extremely flexible; users can ask  almost anything in natural language, and in many cases, receive a reasonable response. However, recent research has demonstrated that even state-of-the-art VLMs from OpenAI and Google consistently underperform as predictors when compared to specialist classifiers, especially on non-traditional modalities and in long-tail domains~\citep{zhang2024visuallygroundedlanguagemodelsbad}. Gemini, GPT-4V and  LLaVA~\citep{liu2023visual} have sought to optimally align language models with specialist embeddings for vision, and in some cases, other modalities as well. However, these efforts have not brought foundation models to parity with specialists --furthermore, for some modalities, such as audio, there is no obvious way to natively utilize a traditional LLM / VLM for predictive tasks.

These challenges motivate our core research question:

\begin{exampleblue}
How can we combine the reasoning capabilities of LLMs with the representational power of specialized models without requiring modality-specific fine-tuning?
\end{exampleblue}

In this work, we posit that visual reasoning, coupled with specialized low-dimensional embedding models, is a skeleton key that unlocks the power of in-context learning and reasoning for arbitrary data modalities and domains.

\begin{example}
\begin{enumerate}
[leftmargin=1em]
\item We propose MARVIS, an efficient, modality-agnostic system for transforming a VLM into a performant predictor. Using a QwenVL model with no specialized reasoning training, MARVIS achieves competitive performance across vision, audio, and tabular modalities, and across a wide range of scientific domains, on both classification and regression tasks.
\item We demonstrate empirically that MARVIS does more than simply copy predictions; it reasons over the available information sources, implicitly analyzing and balancing them to improve its own predictive power. It can rationalize its decisions post-hoc and suggest next steps, unlike the specialist models it adapts.
\item We also introduce numerous valuable secondary contributions to facilitate future research in this area, including the first large-scale standardized tabular classification and regression datasets with complete semantic information (see \cref{sec:semantic-datasets}), a strong FFT baseline for tabular data (see \cref{sec:fft-experiments}), comprehensive ablations, and a well-documented Github repository.\footnote{\url{https://github.com/penfever/marvis}}
\end{enumerate}
\end{example}


\begin{figure*}[!t]
\centering
\adjustbox{trim=0 125pt 0 125pt, clip, width=\textwidth}{%
\includegraphics{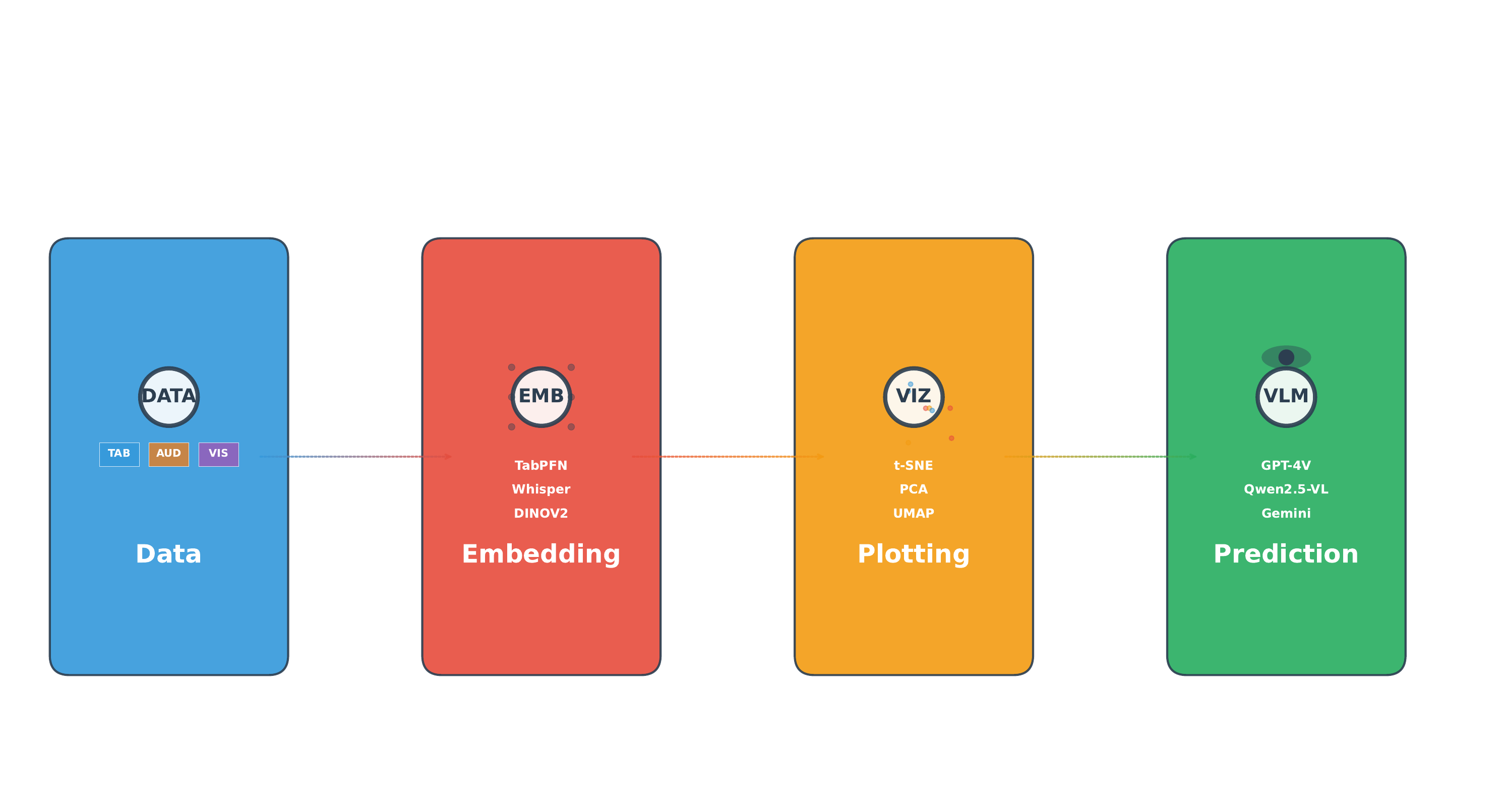}}
\caption{\textbf{The four-stage MARVIS pipeline.} We start with raw input data, capture key patterns using specialist embedding generating models, determine an appropriate strategy for plotting the data, and prompt a VLM with visual context, as well as (optionally) metadata and semantic context, then extract predictions.}
\label{fig:marvis-method}
\end{figure*}

\section{The MARVIS System}
\label{sec:method}

\subsection{Core Insight: Vision as a Skeleton Key}

Relying solely on text to ingest data is limited and does not align with how humans operate. For predictive tasks, it is not usually the raw data that we want the model to reason over; rather, it is a distilled view of that data, for the purposes of answering specific questions or rendering judgments.
Human scholars tend to reason more effectively with data visualizations, simplified views of complex data~\citep{Unwin2020Why,de2025low}. VLMs, which are pretrained on web-scraped data, can understand and interpret a wide range of scientific imagery and visualizations of specialized embedding spaces.
Thus, we posit that \textit{embedding visualizations can operate as skeleton keys}, unlocking any kind of data for vision-language models without requiring modality-specific training beyond vision.
Moreover, visualizations can be easily generated at inference time with standard packages, such as scikit-learn \citep{pedregosa2011scikit}.

\subsection{System Architecture}

MARVIS operates through a four-stage pipeline (\cref{fig:marvis-method}):
\begin{enumerate}
\item \textbf{Embedding Generation}: A domain-appropriate embedding model maps raw input data to vector representations. MARVIS is agnostic to the choice of embedding model; any model that produces fixed-dimensional vectors can be used.
\item \textbf{Dimensionality Reduction}: t-SNE projects the embeddings to 2D, producing a scatter plot visualization optimized for VLM processing. A zoom factor centers the view on the query point and its local neighborhood. The particular dimensionality reduction algorithm is likewise mutable in MARVIS, and we experiment with many.
\item \textbf{Visual Reasoning}: The VLM receives the visualization along with a structured prompt containing the class legend, KNN neighbor analysis, and optional metadata (examples available in \secref{sec:viz-gallery}. It reasons over the spatial layout to produce a prediction, either by class name or color.
\item \textbf{Response Processing}: A parser extracts the predicted class label from the VLM's natural language response using the known class-to-color mapping.
\end{enumerate}

\noindent\textbf{Embedding models.} MARVIS's plug-and-play architecture accepts embeddings from any upstream model. \cref{tab:embedding_models} lists the models used in our experiments; each was selected as a strong representative for its modality. The system requires no modification when switching between modalities---only the embedding model changes.

\begin{table}[h]
\centering
\small
\begin{tabular}{@{}llll@{}}
\toprule
\textbf{Modality} & \textbf{Model} & \textbf{Dim} & \textbf{Training} \\
\midrule
Vision & DINOv2-ViT-L-14 & 1024 & Self-supervised \\
Audio & MS-CLAP & 512 & Contrastive \\
Biology & BioCLIP2 & 512 & Taxonomic contr. \\
Tabular & TabPFNv2 & 256 & In-context learning \\
\bottomrule
\end{tabular}
\caption{\textbf{Embedding models by modality.} MARVIS accepts embeddings from any upstream model; only the embedding model changes between modalities.}
\label{tab:embedding_models}
\end{table}

For vision, DINOv2-ViT-L-14-reg provides robust visual representations trained through self-supervised learning~\citep{oquab2023dinov2}. For audio, Microsoft CLAP creates joint audio-language embeddings via contrastive pre-training~\citep{elizalde2023clap}. For biological data, BioCLIP2 specializes in scientific vision understanding, incorporating taxonomic labels in contrastive training~\citep{stevens2024bioclip}. For tabular data, TabPFNv2 provides embeddings via transformer-based in-context learning~\citep{hollmann_accurate_2025}. Additional details on hyperparameters are provided in \cref{sec:implementation-details}.

\subsection{Visualization Engine}

\begin{figure}[t]
\centering
\includegraphics[page=3, trim=120pt 570pt 120pt 40pt, clip, width=\columnwidth]{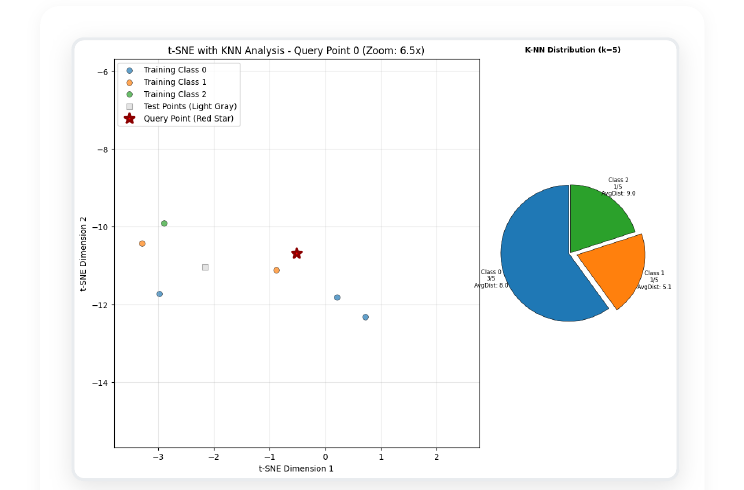}
\caption{\textbf{Example MARVIS input.} A zoomed t-SNE visualization of the CMC dataset as seen by the VLM. The red star marks the query point; colored circles are training points (3 classes); gray squares are test points. The pie chart shows the KNN neighbor distribution computed in the original embedding space. See \cref{sec:viz-gallery} for more examples.}
\label{fig:viz-example}
\end{figure}

The visualization engine transforms high-dimensional embeddings into images that a VLM can interpret. \cref{fig:viz-example} shows an example of what the VLM actually sees at inference time: a zoomed t-SNE scatter plot with color-coded training points, gray test points, a red star marking the query point, and a KNN pie chart summarizing the neighbor distribution in the original embedding space.

\noindent\textbf{Visualization generation.} Even modern VLMs do not ``see'' as well as humans; the particular patch dimensions and the limited range of local attention mean that the VLM performs best when DPI is optimized and a zoom factor is used to enlarge the region of interest around the query point. We find that the optimal zoom factor varies by modality but can be set once per modality (tuning procedure described below). Ideally, the zoom factor is such that the target point and its neighbors are captured within the 14$\times$14 patches from the VLM's sliding window attention, significantly enhancing spatial understanding.

\noindent\textbf{Context composition.} One key design decision is which context to include alongside the visualization. In \cref{sec:ablation-context-choice}, we name and ablate over 25 different configurations, including perturbation-based axes, semantic axes, multi-view layouts, and 3D projections. For the main experiments in this paper, we use the ``tsne\_knn'' setting, which overlays KNN neighbor connections computed in the original high-dimensional embedding space onto the 2D t-SNE plot. This setting offers the best speed/quality tradeoff: because KNN operates on the full embeddings without dimensionality reduction, it is sometimes able to discover relationships that the 2D visualization alone would miss. We find that fixing the nearest neighbors hyperparameter at $\min(30, 10\%\ \text{of the training data})$ works well across dataset sizes and modalities.

\noindent\textbf{Hyperparameter tuning protocol.} All MARVIS hyperparameters except the t-SNE zoom factor are held constant across all experiments at the values listed in \cref{sec:implementation-details} (perplexity $15$, t-SNE iterations $1000$, nearest-neighbor count $\min(30, \lceil 0.1 \cdot |\mathcal{D}_{\mathrm{train}}|\rceil)$, Euclidean metric for general features and cosine for embeddings); no further per-dataset tuning is performed. The zoom factor is the only hyperparameter we tune per setting, and is tuned once per modality rather than per dataset. For each modality we perform a grid search over a small candidate set of zoom values and select the value with highest validation accuracy, using the validation split provided by the benchmark suite for tabular data and a held-out $10\%$ subset of the training data for audio, vision, and biological data. The selected per-modality zoom values are reported in \cref{sec:implementation-details}.

\subsection{VLM Reasoning and Response Parsing}

\noindent\textbf{VLM backend selection.} The choice of VLM architecture is critical: many older architectures either cannot localize what they ``see'' effectively, or cannot ``see'' clearly enough to take advantage of visualizations. We select the 3B parameter Qwen 2.5 VL model~\citep{bai2025qwen25vltechnicalreport} for three reasons:
\begin{enumerate}
\item It uses 14$\times$14 patches with sliding window attention, emphasizing local patch interaction---important for distance-based visualizations where proximity matters.
\item It processes images of arbitrary aspect ratios without distortion, enabling effective multi-visualization layouts.
\item The Qwen 2.5 VL series has been specifically trained on long-context and scientific imagery.
\end{enumerate}

\noindent\textbf{Response parsing.} To avoid the common failure mode in which predictions are correct but not detected by the parser, we enforce consistent color schemes and naming across all visualization legends, ensuring clear visual separation for VLM interpretation. The parser is made aware of both the class names and their assigned colors, and is given an explicit mapping between them. Class names in legends are limited to the classes that actually appear in each visualization, controlling legend size for datasets with many classes. Using this simple modification, we are able to scale MARVIS to a surprising degree; on FishNet, a realistic fine-grained image classification task with 463 classes, MARVIS achieves accuracy only slightly worse than the best specialist.

\subsection{Backend Robustness}
\label{sec:backend_variants}

\begin{figure*}[t]
\centering
\includegraphics[width=0.95\textwidth]{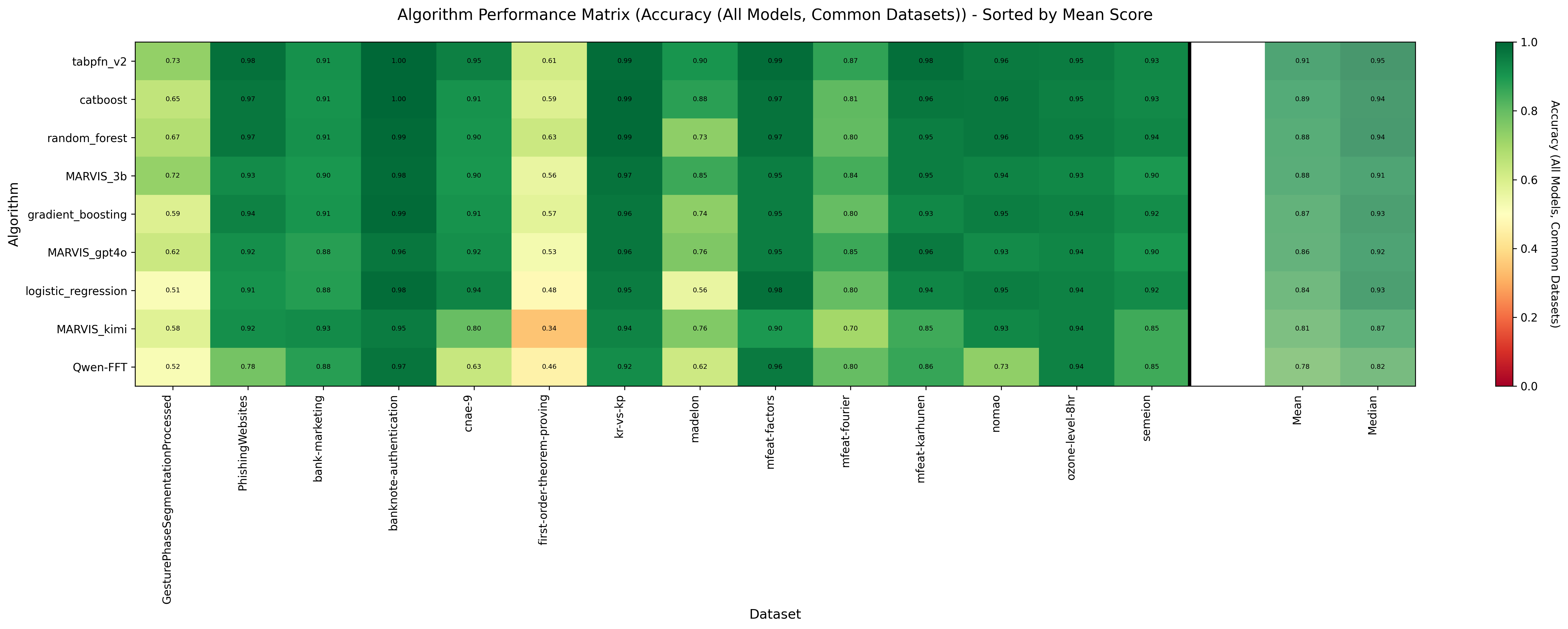}
\caption{\textbf{Accuracy matrix for MARVIS backend variants and FFT.} MARVIS's performance depends considerably more on the choice of embedding model than on the VLM backend: a 3B QwenVL model matches GPT-4o-mini and outperforms a recent thinking model.}
\label{fig:backend_variants}
\end{figure*}

A natural question is whether MARVIS's performance depends critically on the choice of VLM backend. To answer this, we ablate over multiple VLM backends on a subset of the OpenML CC-18 Semantic tabular classification benchmark (\cref{fig:backend_variants}).

The results reveal that MARVIS's performance depends considerably more on the choice of embedding model than on the choice of VLM backend. A small QwenVL 2.5 3B model (MARVIS-3B) matches GPT-4o-mini (MARVIS\_gpt4o) and outperforms a more recent thinking model (Kimi-VL-A3B-Thinking, referenced as MARVIS\_kimi).

\subsection{System Design and Deployment}

\noindent\textbf{Latency breakdown.}
We profile MARVIS using PyTorch device synchronization barriers with \texttt{time.perf\_counter()} for accurate GPU-aware wall-clock timing. \cref{tab:latency} reports per-sample costs averaged over 10 runs on consumer hardware (Apple M-series, MPS backend) with a 200-sample tabular dataset. VLM inference overwhelmingly dominates at 98.9\% of per-sample time. Visualization generation (t-SNE rendering, zoom, KNN overlay, and rasterization) accounts for only 1.1\%, and response parsing is negligible. On server hardware (H100), VLM inference drops to 0.5--2.0s per sample.

\begin{table}[h]
\centering
\small
\begin{tabular}{@{}lrr@{}}
\toprule
\textbf{Stage} & \textbf{Time (ms)} & \textbf{\% Total} \\
\midrule
\multicolumn{3}{@{}l}{\textit{One-time costs (amortized over $N$ test samples)}} \\
\quad Embedding generation & 5702 & --- \\
\quad t-SNE fitting & 752 & --- \\
\midrule
\multicolumn{3}{@{}l}{\textit{Per-sample costs (mean $\pm$ std, $N$=10)}} \\
\quad Visualization generation & $64 \pm 4$ & 1.1\% \\
\quad VLM inference (3B) & $5930 \pm 1143$ & 98.9\% \\
\quad Response parsing & $<$1 & $<$0.1\% \\
\midrule
\textbf{End-to-end per sample} & $\mathbf{5998 \pm 1143}$ & \textbf{100\%} \\
\bottomrule
\end{tabular}
\caption{\textbf{Per-sample latency breakdown} on consumer hardware (Apple MPS, Qwen2.5-VL-3B). VLM inference dominates; all other stages are negligible. On server GPUs (H100), end-to-end time drops to 0.5--2.0s per sample.}
\label{tab:latency}
\end{table}

\noindent\textbf{Parallelism and scalability.}
The MARVIS pipeline admits natural parallelism at multiple levels. Visualization generation is strongly parallel: each test sample's t-SNE plot is independent once the shared t-SNE projection is computed, so $N$ samples can be rendered concurrently across CPU cores. VLM inference can be batched across samples on a single GPU, or distributed across multiple GPUs with standard tensor parallelism. The most expensive one-time costs---embedding generation (5.7s) and t-SNE fitting (0.75s)---are amortized across all test samples; for a 1000-sample test set, these add only 6.5ms per sample. Peak GPU memory remains under 8GB for the 3B model, enabling deployment on consumer hardware (e.g., a single RTX 4090 or Apple M-series laptop).

\noindent\textbf{Modularity.}
MARVIS's four-stage pipeline is fully modular: the embedding model, visualization strategy, VLM backend, and response parser can each be swapped independently. Switching from tabular to audio data requires only changing the embedding model (\cref{tab:embedding_models}); switching from a local 3B model to GPT-4o-mini requires only changing the VLM backend (\cref{fig:backend_variants}); switching from KNN overlays to perturbation axes requires only changing the visualization configuration. No retraining is needed for any swap.

\noindent\textbf{Information available to the VLM.}
The VLM does not receive embedding vectors or raw feature values. Its input, for each test point, is a rasterized 2D scatter plot of the t-SNE projection, a class legend, an optional KNN pie chart summarizing nearest neighbors, and the task prompt. \cref{tab:vlm_inputs} enumerates this interface. This makes MARVIS's VLM interface narrower than approaches that serialize feature values (e.g., TabLLM, JOLT) or transmit embedding vectors.

\begin{table}[h]
\centering
\small
\begin{tabular}{@{}p{0.46\columnwidth}p{0.46\columnwidth}@{}}
\toprule
\textbf{Provided to VLM} & \textbf{Not part of VLM input} \\
\midrule
Rasterized 2D scatter plot & Raw feature values \\
Class legend (labels + colors) & Column names and schema metadata \\
KNN pie chart (neighbor counts) & Original embedding vectors \\
Task prompt & Source row indices \\
\bottomrule
\end{tabular}
\caption{\textbf{Information flow into the VLM.} Left: data the VLM receives for each test point. Right: data produced or consumed by MARVIS that is not part of the VLM input.}
\label{tab:vlm_inputs}
\end{table}

The interface is narrow but structured. From the plot and legend the VLM can infer---and must, for its reasoning to be useful---the number and relative sizes of classes; the neighborhood structure of classes in the embedding space (which classes cluster together, which overlap, where a test point sits relative to them); and the presence of clusters, modes, and outliers. When semantic class names are used, the class taxonomy of the task is also exposed. When semantic-axes or perturbation-axes contexts are enabled, named feature directions and their local effects are exposed through plot annotations and the prompt, effectively revealing part of the feature schema.

For completeness we note two further, more conservative properties of these inputs. First, the plot and legend are in principle consistent with membership inference~\citep{shokri2017membership}: a party holding the embedding model and a candidate raw input can forward-project the candidate and compare its location against the plot, which may reveal whether a similar point was present in the visualized dataset. Second, because outliers sit visibly apart from clusters, atypical examples in the dataset are individually more distinguishable in the plot than typical examples are.

The default MARVIS configuration used throughout our experiments (t-SNE scatter plot with KNN overlay) does not transmit semantic class names, named feature directions, or raw feature values; the class legend identifies classes by index only. Semantic class names, semantic axes, and perturbation axes are optional context-composition variants, explored in \cref{fig:mean_acc_by_config}, that widen this interface beyond the default.

\section{Experiments}

\noindent\textbf{Overview.} Our main experiments assess MARVIS across four distinct modalities using domain-appropriate embedding models and established benchmarks; we compare against both specialized baselines and alternative LLM/VLM approaches.

\cref{tab:comprehensive_results} presents MARVIS performance across all modalities compared to 5 specialized baselines and 4 alternative LLM/VLM approaches. For each benchmark, we conduct a single MARVIS run. We use a QwenVL 2.5 3B Instruct backbone. MARVIS hyperparameters are set as described in \cref{sec:method}: the t-SNE zoom factor is selected per modality by grid search on a held-out validation split; all other hyperparameters are fixed across benchmarks. The LLM / VLM baseline results in the paper are reported using the best performing LLM / VLM in the class (we consider QwenVL 2.5 3B Instruct and Gemini-Flash-2.0 via the Gemini API). All MARVIS results are zero-shot in the sense that we do not give examples of the task to the VLM at inference time; they are full-shot in the sense that the embedding-generating models have access to the entire test set without labels. For the LLM / VLM baselines, image classification is performed zero-shot. Tabular classification and regression uses  the JOLT~\citep{shysheya2025joltjointprobabilisticpredictions} and TabLLM~\citep{hegselmann2023tabllm} strategies with k-shot computed dynamically based on the maximum context length. We report the best result in the table. Specialist models are full-shot, and we report the best overall result in the table. For extended results, a detailed description of the method we use to generate our novel tabular benchmarks CC18-Semantic and Regression2025-Semantic,  and a deeper dive into tabular data, including balanced metrics, please refer to \cref{sec:tabularanalysis}.

\noindent\textbf{Specialized model baselines.} For vision, the best performing specialist was the large DinoV2 model with a registry and KNN classification~\citep{oquab2023dinov2}. For audio, the CLAP model with contrastive zero-shot classification from Microsoft and OpenAI's Whisper-V2-Large model with KNN classification perform the best~\citep{radford2022robustspeechrecognitionlargescale,elizalde2023clap,ma-etal-2024-investigating}. For biological data, BioCLIPv2 with KNN classification performs the best~\citep{gu2025bioclip2emergentproperties}. For tabular data, TabPFNv2 with standard forward pass classification and regression is a strong baseline; we also consider classical baselines such as CatBoost and linear models in \cref{sec:tabularanalysis}~\citep{prokhorenkova2018catboost,hollmann_accurate_2025}.

\begin{table*}[!t]
\centering
\caption{\textbf{Domain-specific embeddings, benchmarks, and detailed results.}
Results are boldfaced when statistically tied for best performance within 95\% confidence intervals (normal approximation).
MARVIS demonstrates competitive or superior performance on most individual
benchmarks, achieving average results within 2.5\% of an ensemble of specialized methods while providing universal applicability. Benchmark acronyms: C10 = CIFAR-10, C100 = CIFAR-100, ESC =
ESC-50, RAV = RAVDESS, US8 = UrbanSound8K, FSH = FishNet, AWA = AWA2, PLD =
PlantDoc, CC18 = OpenML CC18, R25 = Regression 2025.
We show the best results of specialized models and traditional LLM/VLM approaches. For all benchmarks except R25, the metric is Accuracy. For R25, it is R2 Score (with a minimum score of 0). The number reported is the mean over all sub-tasks for multi-task benchmarks.}
\label{tab:comprehensive_results}
\small
\begin{tabular}{@{}ll l c c c c r@{}}
\toprule
Domain & Embeddings & Bench. & Size (K) & MARVIS & Specialist & LLM/VLM & 95\% CI \\
\midrule
\multirow{2}{*}{Vision} & \multirow{2}{*}{DINOV2} & C10 & 60 & 98.0 & \textbf{99.0} (DINOV2) & 85.7 (Gemini) & $\pm$0.1 \\
& & C100 & 60 & 88.0 & \textbf{91.6} (DINOV2) & 64.3 (Gemini) & $\pm$0.3 \\
\midrule
\multirow{3}{*}{Audio} & \multirow{3}{*}{CLAP} & ESC & 2 & \textbf{91.3} & \textbf{90.5} (CLAP) & -- & $\pm$1.2 \\
& & RAV & 1.4 & 38.4 & \textbf{47.9} (Whisper) & -- & $\pm$2.5 \\
& & US8 & 8.7 & \textbf{79.8} & 77.1 (CLAP) & -- & $\pm$0.8 \\
\midrule
\multirow{3}{*}{Biological} & \multirow{3}{*}{BioCLIP2} & FSH & 94 & 80.2 & \textbf{83.7} (BioCLIP) & 59.5 (Gemini) & $\pm$0.3 \\
& & AWA & 37 & 95.7 & \textbf{97.1} (BioCLIP) & 96.5 (Gemini) & $\pm$0.2 \\
& & PLD & 2.5 & 67.4 & \textbf{72.0} (BioCLIP) & \textbf{74.2} (Gemini) & $\pm$1.8 \\
\midrule
\multirow{2}{*}{Tabular} & \multirow{2}{*}{TabPFNv2} & CC18 & 155 & 84.5 & \textbf{87.8} (TabPFNv2) & 50.1 (TabLLM) & $\pm$0.2 \\
& & R25 & 35 & 66.0 & \textbf{67.0} (TabPFNv2) & 05.1 (JOLT) & $\pm$0.5 \\
\midrule
\multicolumn{3}{@{}l}{\textbf{(Score, \# Models)}} & -- & (78.9, 1) & (81.4, 5) & (62.2, 4) & -- \\
\bottomrule
\end{tabular}
\end{table*}

\noindent\textbf{LLM / VLM baselines.} For vision, we use the standard strategy of zero-shot prompting and exact match extraction described in works such as \citep{zhang2024visuallygroundedlanguagemodelsbad}. For audio, we are unable to compare to public API-based models, as to the best of our knowledge, no generalist exists capable of performing audio classification.

\noindent\textbf{LLM tabular baselines.} In the tabular domain, as a secondary contribution, we generate the first large-scale standardized benchmarks for tabular classification and regression that include semantic class names, feature names and metadata; CC18-Semantic and Regression 2025 Semantic. We also re-implement two prominent LLM-tabular methods, TabLLM and JOLT~\citep{hegselmann2023tabllm,shysheya2025joltjointprobabilisticpredictions}, which lack general-purpose implementations. For more details on this, please refer to \cref{sec:tabularanalysis}.

\noindent\textbf{Additional details.} For more analysis on the embedding models and baselines, please refer to \cref{sec:implementation-details}. For more explanation of the benchmarks we use, please refer to \cref{sec:benchmark-descr}.

\subsection{Findings}

\textbf{MARVIS is competitive with SOTA specialist predictors. } Across a wide range of modalities, we observe that MARVIS strongly conserves predictive performance -- across most tasks we consider, it is able to match the best specialist model in the cohort. By comparison, the best existing LLM / VLM methods, tailored for each domain, achieve ~77\% of specialist performance on average. Remarkably, we find that MARVIS is a more accurate image classifier than Gemini Flash 2.0, despite never actually having seen the images. MARVIS also sometimes improves on specialists; it outperforms CLAP, a specialist contrastive predictor, using its own embeddings.

\begin{example}
MARVIS-3B achieves competitive performance across four distinct modalities, approaching and occasionally exceeding the best specialist predictors, and improving on LLM / VLM-only methods by 16.7\%.
\end{example}

\textbf{MARVIS outperforms direct fine-tuning of its base model. } In \cref{sec:fft-experiments}, we describe a novel method for fine-tuning an LLM directly on the embeddings of an upstream model such as TabPFNv2. We test this method (Qwen-FFT) at inference time and find that it is highly accurate, far outperforming previously published strategies such as JOLT and TabLLM for general-case tabular inference with LLMs; however, as shown in \cref{sec:backend_variants}, MARVIS-3B outperforms even this strong baseline on average. This result may be surprising -- although it is time-consuming and costly, full fine-tuning is generally considered the gold standard for accuracy. However, we find that, although the FFT solution generally reduces training loss to near-zero, it can fail to generalize, particularly when the training dataset size is small. This finding suggests that the visual reasoning paradigm of MARVIS is a more robust approach to leveraging VLMs for prediction than direct fine-tuning on embeddings.

\begin{figure}[h]
\centering
\includegraphics[width=0.95\linewidth]{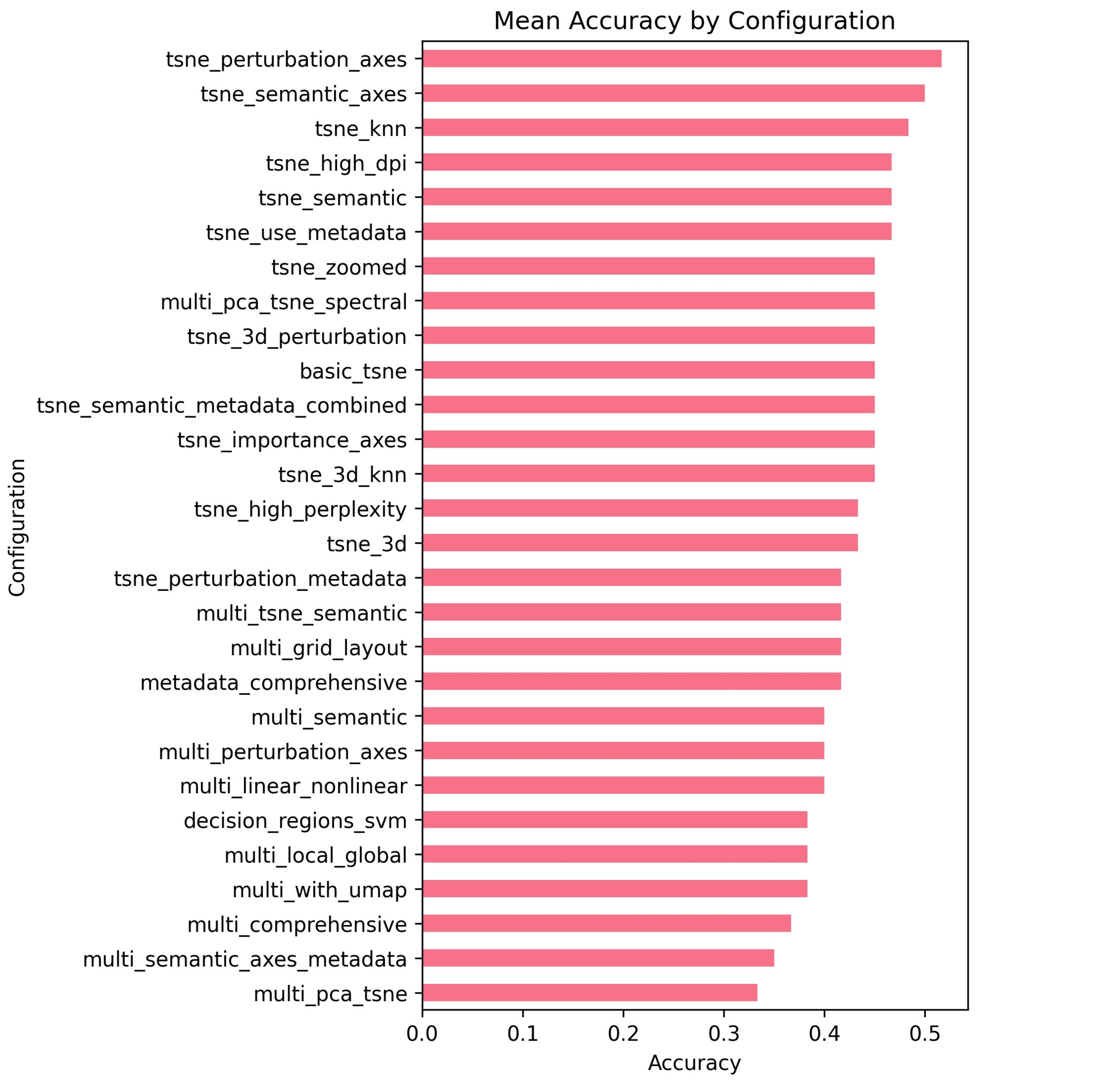}
\caption{\textbf{The selection of context strongly influences MARVIS performance.} We ablate over twenty different context composition strategies, and find that perturbation-based approaches with uncertainty analysis achieve the highest performance, followed by semantic axes with meaningful class labels. The majority of the experiments in the paper are conducted using TSNe + KNN, which we adopt as our default configuration.}
\label{fig:mean_acc_by_config}
\end{figure}

\textbf{VLMs reason over their input data and condition their behavior based on the context provided.} One core research question, from our perspective, was whether a VLM was simply copying learned patterns or utilizing simple heuristics to achieve this strong performance. Systematic analysis of VLM reasoning in \cref{fig:mean_acc_by_config} demonstrates clear correlations between reasoning quality and metric gains, on average, across three tabular classification datasets (two with meaningful semantic features, one without).

Further analysis of disagreement patterns reveals that only 35\% of methods agree on all test cases, with 65\% showing partial disagreement. Furthermore, in \cref{tab:method_signatures}, we show that different visualization methods elicit systematically different reasoning approaches, providing strong evidence that VLMs adapt their analysis based on visual information content. Still more evidence can be found in \cref{sec:reasoning_pattern_analysis}.
\begin{sloppypar}
We observe that different visualization methods elicit systematically different reasoning approaches, providing strong evidence that VLMs adapt their analysis based on the available visual information. \textbf{tsne\_knn} produces quantitative neighbor analysis with explicit distance calculations (average 48.0 words), \textbf{tsne\_semantic\_axes} integrates semantic class information with spatial reasoning (304.9 character responses) and \textbf{tsne\_perturbation\_axes}  generates the longest, most detailed responses (310.6 characters) with sophisticated uncertainty analysis.
\end{sloppypar}

These patterns suggest that VLMs engage in more thorough spatial analysis when the visual information supports accurate classification, indicating genuine reasoning rather than pattern matching.

\begin{table}[h]
\centering
\caption{\textbf{Method-Specific Reasoning Patterns.} Each visualization method elicits distinct reasoning behaviors: k-NN methods trigger quantitative distance analysis, perturbation methods generate longer responses, and basic methods rely heavily on proximity heuristics. Here, Resp. Length refers to the token count of responses, distance mentions to the rate at which the response mentions distance between points in embedded space, and closest usage refers to how often MARVIS uses the word ``closest'' in its response.}
\label{tab:method_signatures}
\resizebox{\columnwidth}{!}{%
\begin{tabular}{l|c|c|c}
\toprule
\textbf{Method} & \textbf{Resp. Length} & \textbf{Dist. Mentions} & \textbf{Closest Usage} \\
\midrule
tsne\_3d\_perturbation & \textbf{365.3} & 0.000 & 0.433 \\
tsne\_perturbation\_axes & 310.6 & 0.000 & \textbf{0.650} \\
tsne\_semantic\_axes & 304.9 & 0.000 & 0.683 \\
tsne\_knn & 279.0 & \textbf{0.650} & 0.883 \\
basic\_tsne & 268.3 & 0.000 & 1.000 \\
\bottomrule
\end{tabular}%
}
\end{table}

\textbf{Reasoning quality predicts classification accuracy.} Beyond method-level differences, we find that \textit{individual} prediction quality correlates with measurable reasoning features. \cref{tab:reasoning_quality} analyzes 83 experimental configurations across multiple test cases, comparing the reasoning traces of correct vs.\ incorrect predictions. Correct predictions exhibit longer, more detailed responses with more color mentions and explicit distance reasoning. Critically, incorrect predictions rely significantly more on simple heuristics---words like ``closest'' (+0.21) and ``majority'' (+0.20)---suggesting that when the VLM falls back on shallow shortcuts instead of engaging with the spatial structure, accuracy suffers.

\begin{table}[h]
\centering
\caption{\textbf{Reasoning Quality Predicts Accuracy.} Correct predictions exhibit longer, more sophisticated responses with increased spatial analysis and reduced reliance on simple heuristics, across 83 experimental configurations.}
\label{tab:reasoning_quality}
\resizebox{\columnwidth}{!}{%
\begin{tabular}{l|c|c|c}
\toprule
\textbf{Reasoning Feature} & \textbf{Correct} & \textbf{Incorrect} & \textbf{$\Delta$} \\
\midrule
Response Length (chars) & 281.2 & 268.3 & \textbf{+12.9} \\
Color Mentions & 1.85 & 1.52 & \textbf{+0.33} \\
Distance Reasoning & 0.074 & 0.057 & \textbf{+0.018} \\
\midrule
``Closest'' Heuristic & 0.56 & 0.77 & \textbf{$-$0.21} \\
``Majority'' Heuristic & 0.05 & 0.25 & \textbf{$-$0.20} \\
\bottomrule
\end{tabular}%
}
\end{table}

\begin{sloppypar}
This pattern provides concrete evidence that MARVIS accuracy is driven by genuine visual reasoning. The following excerpts from VLM reasoning traces illustrate how different visualization methods elicit qualitatively different reasoning strategies:
\end{sloppypar}

\smallskip
\noindent\textit{With k-NN overlay} (quantitative distance reasoning):
\begin{quote}
\small ``The query point is closer to the cluster of Class\_1 neighbors (4 neighbors) than to the cluster of Class\_2 neighbors (1 neighbor). Additionally, the average distance to Class\_1 neighbors (6.1) is slightly lower than to Class\_2 neighbors (5.2), indicating higher similarity to Class\_1.''
\end{quote}

\noindent\textit{With semantic axes} (class-label integration):
\begin{quote}
\small ``The red star (query point) is closest to the orange-colored points, which represent the `Long-term methods' class. This spatial clustering indicates that the query point is more aligned with the characteristics of the `Long-term methods' class.''
\end{quote}

\noindent\textit{Basic t-SNE only} (proximity heuristic):
\begin{quote}
\small ``The red star (query point) is closest to the green-colored training points, which are associated with Class\_2.''
\end{quote}

\smallskip
\noindent Quantitative distance calculations appear \textit{only} when k-NN information is provided; semantic reasoning emerges \textit{only} with meaningful class labels; and without either, the VLM falls back on simple proximity. This adaptive behavior---not fixed pattern matching---explains why richer visualization contexts yield higher accuracy (\cref{fig:mean_acc_by_config}). For additional analysis, see \cref{sec:vlm_reasoning_analysis}.

\textbf{MARVIS performance is robust under an inductive-only t-SNE projection.} The default MARVIS pipeline fits t-SNE jointly on training and test embeddings (transductive). To test whether MARVIS could be deployed inductively---fitting t-SNE on the training set alone and projecting each test point in afterward, one at a time---we replace the transductive step with an explicit out-of-sample projector. t-SNE has no native out-of-sample extension (sklearn's \texttt{TSNE} provides only \texttt{fit\_transform}, with no learned function to apply to a new point), so we deliberately choose the simplest plausible parametric approximation: a \emph{linear regression} $f(x) = Wx + b$, fit by closed-form least squares from each training embedding $x_i$ to its training-only t-SNE coordinate $y_i$. At inference, each test point is projected independently as $\hat{y}_{\text{test}} = f(x_{\text{test}})$. We use this cheap baseline rather than a more sophisticated method (such as parametric t-SNE or per-point optimization against a frozen training layout) precisely because any stronger inductive projector should perform at least as well; the linear regression therefore acts as a lower bound on what an inductive MARVIS deployment could achieve. All other components are unchanged from the main paper: DINOv2-ViT-L/14-reg, $k=30$ KNN overlay computed in the original embedding space (so KNN statistics are identical across conditions), zoom factor 15, Qwen2.5-VL-3B. \cref{tab:inductive_ablation} reports accuracy on 1{,}000 held-out CIFAR-10 test images. The inductive variant retains 96.4\% accuracy, a drop of only 1.4 percentage points relative to the transductive baseline, and per-sample VLM latency is unchanged. Even this deliberately weak projector preserves nearly all of the model's classification accuracy, indicating that MARVIS's performance is largely determined by the training-set t-SNE map and the VLM's reasoning over it, rather than by transductive optimization with test points in scope.

\begin{table}[h]
\centering
\caption{\textbf{Inductive-only t-SNE ablation on CIFAR-10.} Replacing transductive joint t-SNE with a train-only fit plus linear inductive projection of test points reduces accuracy by 1.4 percentage points at matched compute. Both rows: 50{,}000 training images, 1{,}000 test images, DINOv2-ViT-L/14-reg, $k=30$, zoom factor $=15$, Qwen2.5-VL-3B.}
\label{tab:inductive_ablation}
\small
\begin{tabular}{lccc}
\toprule
\textbf{Variant} & \textbf{Accuracy} & \textbf{t-SNE fit} & \textbf{VLM inf.} \\
\midrule
Transductive (default) & 97.8\% & 212s & 4076s \\
Inductive (LR)         & 96.4\% & 169s & 4009s \\
\bottomrule
\end{tabular}
\end{table}

\begin{figure*}[!t]
\centering
\includegraphics[width=\textwidth]{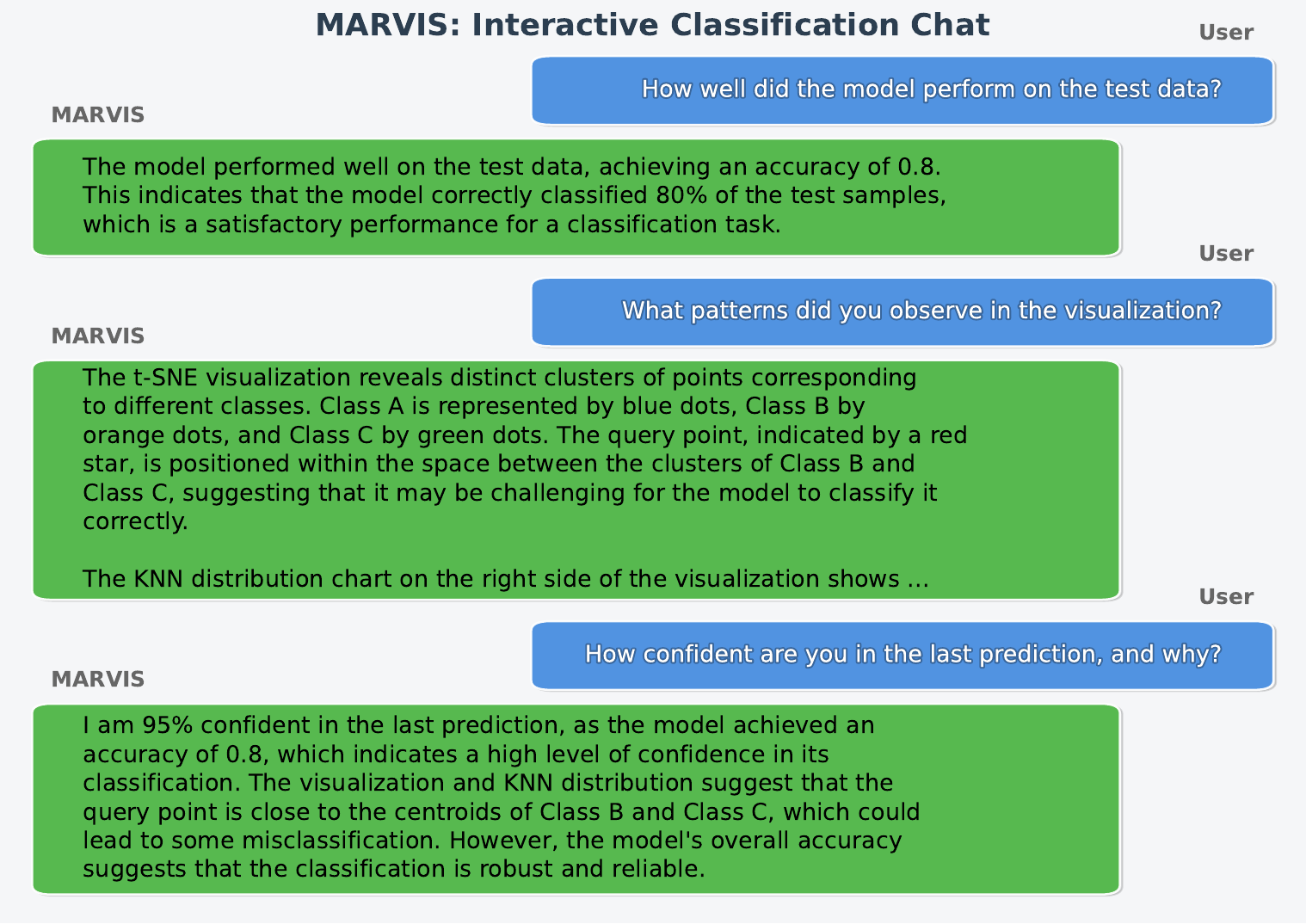}
\caption{\textbf{MARVIS extends traditional predictive capabilities.} Because it requires no fine-tuning, and because it exposes the VLM's classification process to the VLM itself, MARVIS enables VLMs to reason over, and converse about, their predictive performance.}
\label{fig:marvis_chat}
\end{figure*}

\textbf{The flexibility of MARVIS allows for more complex use cases. } In \cref{fig:marvis_chat}, we demonstrate one such use case -- open-ended chat about a particular predictive result. In this example, the user asks MARVIS to assess its own performance and recommend strategies to improve results in the future.

\section{Related Work}

MARVIS builds on extensive prior work in vision-language models (VLMs) which has followed two primary evolutionary tracks: maximalist approaches from industry labs focusing on peak performance, and minimalist open-source approaches prioritizing efficiency and accessibility; in \cref{sec:extendedrelatedworks}, we trace the history of this evolution in greater detail.

The use of embedding spaces for cross-modal understanding has roots in representation learning~\citep{bengio2013representation} and dimensionality reduction techniques~\citep{van2008visualizing}. Recent work has explored the geometric properties of embedding spaces~\citep{ethayarajh2019contextual} and their visualization for interpretability~\citep{liu2017towards}. t-SNE and UMAP have been widely used for visualizing high-dimensional data~\citep{mcinnes2018umap}, but their application to VLM reasoning represents a novel paradigm. Previous work on visual reasoning has focused on spatial relationships in natural images~\citep{johnson2017clevr}, but MARVIS extends this to abstract embedding spaces across arbitrary modalities.

MARVIS distinguishes itself from existing approaches through several key innovations: (1) \textbf{Training-free adaptation}: Unlike approaches requiring extensive fine-tuning, MARVIS leverages pre-trained components without modification; (2) \textbf{Universal modality support}: A single architecture handles any data type through embedding visualization; (3) \textbf{Computational efficiency}: Achieves competitive performance with a 3B parameter model versus much larger specialized systems.

\section{Conclusion}

We introduce MARVIS, a method that enables small VLMs to predict across any data modality through embedding visualization. By transforming embedding spaces into visual representations optimized for VLM spatial reasoning, MARVIS achieves competitive performance across diverse domains.

MARVIS addresses key limitations in existing approaches: it requires no domain-specific training and maintains competitive performance across modalities. The approach demonstrates that visual reasoning can serve as a universal interface for foundation models across any data modality.

Based on this, we propose several key principles for designing effective VLM interfaces:

\begin{itemize}
\item \textbf{Information density matters}: Richer visualizations elicit more sophisticated reasoning
\item \textbf{Method-purpose alignment}: Different visualization approaches suit different reasoning tasks
\item \textbf{Adaptive interface design}: VLMs can effectively utilize different types of visual information
\end{itemize}

Future work includes further investigation of the optimal mix of visualizations and embeddings to boost performance and fine-tuning strategies which may improve the performance of base VLMs for reasoning over scientific imagery, including reasoning post-training.

\bibliographystyle{ACM-Reference-Format}
\bibliography{marvis}

\appendix
\sloppy

\clearpage


\tableofcontents
\newpage

\section{APPENDIX: Benchmark Dataset Descriptions}
\label{sec:benchmark-descr}

\subsection{Vision Benchmarks}

\textbf{CIFAR-10}:  One of the most widely used datasets for computer vision research: contains 60,000 32×32 color images in 10 classes (airplanes, cars, birds, cats, deer, dogs, frogs, horses, ships, trucks) with 6,000 images per class. Split into 50,000 training and 10,000 test images~\cite{Krizhevsky2009LearningML}.

\textbf{CIFAR-100}: Similar to CIFAR-10 but with 100 classes containing 600 images each (500 training, 100 test per class). The 100 classes are grouped into 20 superclasses, making this a more challenging classification benchmark.

\subsection{Audio Benchmarks}

\textbf{ESC-50 (Environmental Sound Classification)}: Contains 2,000 environmental audio recordings with 50 classes and 40 clips per class. Each clip is 5 seconds long at 44.1 kHz, single channel, extracted from public field recordings through Freesound.org~\cite{piczak2015dataset}.

\textbf{RAVDESS (Ryerson Audio-Visual Database of Emotional Speech and Song)}: Audio dataset focusing on emotion recognition tasks, commonly used for evaluating emotional speech and song recognition capabilities~\cite{livingstone_ryerson_2018}.

\textbf{UrbanSound8K}: Contains 8,732 labeled sound excerpts with 10 classes of outdoor/urban sounds, specifically designed for benchmarking sound classification models in urban environments.

\subsection{Biological/Scientific Vision Benchmarks}

\textbf{FishNet}: Large-scale dataset with 94,532 images from 17,357 aquatic species, organized by biological taxonomy (8 classes, 83 orders, 463 families, 3,826 genera). Includes bounding box annotations and supports classification, detection, and functional trait prediction tasks~\cite{khan_fishnet_2023}. We treat FishNet as a classification problem over families.

\textbf{AWA2 (Animals with Attributes 2)}: Animal classification dataset used for zero-shot learning tasks, focusing on learning representations with animal attributes. Part of challenging benchmarks alongside CUB and SUN datasets~\cite{awa2}. We treat AWA2 as a 50-class classification problem with no holdout classes.

\textbf{PlantDoc}: Contains 2,569 images across 13 plant species and 30 classes (diseased and healthy) with 8,851 total labels. Split into 2,328 training and 237 test images, with unbalanced classes ranging from 50-180 images per class~\cite{plantdoc}.

\subsection{Tabular Benchmarks}

\textbf{OpenML CC18}: Curated benchmark suite of 72 classification datasets from OpenML 69 of which we utilize), selected based on strict criteria:
\begin{itemize}
\item Size: 500-100,000 observations, $\leq$ 5,000 features
\item Quality: No artificial data, minority/majority class ratio $\geq$0.05
\item Usability: Compatible with multiple algorithms, representing commonly used ML datasets
\end{itemize}

See \cite{bischl2021openmlbenchmarkingsuites} for more on this benchmark, including the complete specification of tasks.

\textbf{Regression 2025}: Custom benchmark of 43 regression tasks from 2015-2025 sourced from OpenML, evaluated using R² scores on a 0-100 scale for consistent comparison across tasks; introduced onto the OpenML platform in March 2025.\footnote{\url{https://openml.org/search?type=benchmark&sort=tasks_included&study_type=task&id=455}} Please follow the link for the complete list and specification of tasks. After discarding tasks on which all models fail, we compute our scores on a subset of 33.

\section{Implementation Details}
\label{sec:implementation-details}

This section contains additional experimental details from the paper.

\subsection{Embedding Models}
\label{sec:embedding-models}

See \cref{tab:embedding_models} in the main body for an overview of the embedding models used per modality. Here we provide additional context on the biological embedding model lineage: BioCLIP2 is the latest in a series of foundation models for biological applications, initiated by BioCLIP, which incorporated taxonomic labels in the vision-language contrastive training~\cite{stevens2024bioclip}.
Follow-up work scaled data to $162$M images~\cite[BioTrove,][]{yang2024biotrove}, specialized the data to camera traps~\cite[CATALOG and WildCLIP,][]{gabeff2024wildclip,santamaria2025catalog}, and added additional model modalities~\cite[TaxaBind,][]{sastry2025taxabind}. For tabular data, TabPFN~\cite{hollmann2022tabpfn} employed transformer-based in-context learning, and was later extended to support larger datasets~\cite{feuer2024tunetablescontextoptimizationscalable,hollmann_accurate_2025,müller2025mothernetfasttraininginference}.

\subsection{Hyperparameters}

In this section, we document the hyperparameters used for our main experiments section.

\textbf{t-SNE Configuration}: 
\begin{itemize}
\item Perplexity: 15 (optimized through ablation studies)
\item Iterations: 1000 for stable convergence
\item Learning rate: 200 (default)
\item Random state: Fixed for reproducibility
\end{itemize}

\textbf{KNN Configuration}
\begin{itemize}
\item nn = 30
\item metric = 'euclidean' (general), 'cosine' (embeddings)
\item weights = 'distance'
\end{itemize}

\textbf{Zoom Factor (per-modality).} Selected by grid search on validation data as described in \cref{sec:method}.
\begin{itemize}
\item Tabular (classification and regression): 2.0
\item Audio: 7.0
\item Vision: 15.0
\item Biological: 20.0
\end{itemize}

\textbf{Tabular Baseline Models Configuration}:

\textbf{CatBoost (Classification \& Regression)}
\begin{itemize}
\item iterations: 1000
\item depth: 6
\item learning\_rate: 0.03
\item random\_seed: 42
\item verbose: False
\item Categorical features: Auto-detected and preserved
\end{itemize}

\textbf{TabPFN v2 (Classification \& Regression)}
\begin{itemize}
\item n\_estimators: 8
\item device: Auto-detected (CUDA if available)
\item ignore\_pretraining\_limits: True
\item Target preprocessing: Quantile binning for regression
\item Max quantiles: min(n\_samples // 2, 1000)
\item NaN/INF imputation: Median strategy
\end{itemize}

\textbf{Random Forest (Classification \& Regression)}
\begin{itemize}
\item n\_estimators: 100
\item max\_depth: None (unlimited)
\item random\_state: 42
\item n\_jobs: -1 (all cores)
\end{itemize}

\textbf{Gradient Boosting (Classification \& Regression)}
\begin{itemize}
\item n\_estimators: 100
\item learning\_rate: 0.1
\item random\_state: 42
\item Feature selection: Max 500 features (SelectKBest)
\end{itemize}

\textbf{Logistic/Linear Regression}
\begin{itemize}
\item max\_iter: 1000 (Logistic only)
\item C: 1.0 (Logistic regularization)
\item random\_state: 42
\item n\_jobs: -1 (all cores)
\item Preprocessing: StandardScaler applied
\end{itemize}

\section{Computational Efficiency}

See \cref{tab:latency} in the main body for a detailed per-sample latency breakdown and discussion of parallelism and modularity.

\textbf{GPU Utilization}: For development and testing combined, we estimate 1,500 H100-hours were used during the creation of this paper. All experiments were conducted on 1$\times$H100 80GB GPUs on a hosted Lambda cluster.

\section{Full Finetuning Experiments}
\label{sec:fft-experiments}

As a strong baseline for MARVIS, we introduce a novel approach to LLM fine-tuning, projecting a sequence of positionally encoded TabPFNv2 embeddings and learned label tokens into the model's token space. At inference time, we project the test element embedding from TabPFNv2 into the model's token space and conduct standard autoregressive inference to acquire the predicted label.

\subsection{Balanced Prefix Construction}
We construct a balanced, few-shot prefix from training embeddings using \texttt{prepare\_tabpfn\_embeddings\_for\_prefix}. Given class labels $y$ and train embeddings $E \in \mathbb{R}^{N \times d}$ (after robust scaling and optional resizing), we select a total of \texttt{num\_few\_shot\_examples} examples across classes, distributing as evenly as possible; short classes are repeated to meet demand. The resulting prefix tensor $P \in \mathbb{R}^{M \times d}$ (with class labels $c \in \{0,\dots,K-1\}^M$) is saved to \texttt{prefix\_data.npz}.

\subsection{Special Tokens and Class Tokens}
We extend the tokenizer with two sentinel tokens \verb|<PREFIX_START>| and \verb|<PREFIX_END>| and with up to 10 class tokens \verb|<CLASS_i>|. The underlying embedding matrix is resized accordingly. These token IDs delimit the region where external embeddings will be injected and provide stable referents for class-conditional evidence tokens.

\subsection{Position-wise Projection into Token Space}
\paragraph{Implementation.} The core mechanism is implemented via \texttt{QwenWithPrefixEmbedding}:
\begin{itemize}[leftmargin=*,nosep]
  \item A learnable projector is defined as \verb|Linear(d, H)|, mapping TabPFNv2 embedding dimension $d$ to the LLM hidden size $H$.
  \item During \verb|forward|, we build \verb|inputs_embeds| from \verb|input_ids| and locate the span between \verb|<PREFIX_START>| and \verb|<PREFIX_END>|. Let the number of available positions be $T$.
  \item If embeddings and class labels are provided, we compute $\tilde P = P W + b \in \mathbb{R}^{M \times H}$ and interleave with class token embeddings: even positions receive projected vectors, odd positions the embeddings of \verb|<CLASS_{c_j}>|, truncated to $T$.
  \item If only embeddings are provided, we fill the $T$ positions with $\tilde P$ contiguously.
  \item The modified \verb|inputs_embeds| are passed to the base model with \verb|input_ids=None|.
\end{itemize}

\paragraph{Rationale and soundness.}
\begin{enumerate}[leftmargin=*,nosep]
  \item \textbf{Representation Alignment.} A learned affine map is the minimal adapter aligning TabPFN geometry to the LLM token manifold, akin to prefix/prompt-tuning adapters.
  \item \textbf{Token-Sequential Semantics.} Injecting a bounded token span leverages positional mixing and attention for fusion with the downstream textual prompt; class-token interleaving ties directions in $\tilde P$ to discrete label anchors.
  \item \textbf{Identifiability.} With only the projector and last $k$ layers unfrozen, gradients supervise a compact subspace, preserving language priors while enabling consistent task adaptation. Another parameter-efficient approach which we do not consider in this draft, LORA, would likely produce similar outcomes.
\end{enumerate}

\subsection{Backbone and Hooks}
The default backbone is \texttt{Qwen/Qwen2.5-3B-Instruct} (configurable via \texttt{--model\_id}). MARVIS prepares the model with prefix-embedding tokens and class tokens using \texttt{prepare\_qwen\_with\_prefix\_embedding}. Optional Vector Quantization (VQ) is available via \texttt{prepare\_qwen\_with\_vq\_prefix\_embedding}.

\subsection{Label Encoding}
We encode labels with a \texttt{LabelEncoder} fitted on train+val+test labels per task; IDs index into the class token set. For float labels near-integral, we cast to integers; otherwise, regression handling is separate.

\begin{figure*}[t]
\centering
\resizebox{\textwidth}{!}{%
\begin{tikzpicture}[
    node distance=6mm and 10mm,
    box/.style={draw, rounded corners, align=center, fill=gray!5, inner sep=3pt, text width=2.8cm},
    proc/.style={draw, rounded corners, align=center, fill=green!5, inner sep=3pt, text width=2.8cm},
    io/.style={draw, align=center, fill=blue!5, inner sep=3pt, text width=2.8cm},
    token/.style={draw, align=center, fill=purple!8, inner sep=2pt, text width=2.8cm},
    every node/.style={font=\small},
    >=Latex
]
    \node[io] (p) {TabPFNv2\\embeddings $P\in\mathbb{R}^{M\times d}$};
    \node[proc, right=of p] (proj) {Linear projector\\$W\!:\,\mathbb{R}^d\!\to\!\mathbb{R}^H$};
    \node[box, right=of proj] (ptilde) {Projected\\$\tilde P\in\mathbb{R}^{M\times H}$};
    \node[token, above=of ptilde] (classes) {Class token\\embeddings $E_c$};
    \node[box, right=of ptilde, xshift=4mm] (inter) {Interleave $[\tilde p_1,E_{c_1},\tilde p_2,E_{c_2},\dots]$};
    \node[token, right=of inter, xshift=4mm] (span) {Insert between\\$<\!PREFIX\_START\!>$ and $<\!PREFIX\_END\!>$};
    \node[proc, right=of span, xshift=4mm] (qwen) {Qwen blocks\\(self-attn over span + prompt)};
    \draw[->] (p) -- (proj);
    \draw[->] (proj) -- (ptilde);
    \draw[->] (ptilde) -- (inter);
    \draw[->] (classes) -- (inter);
    \draw[->] (inter) -- (span);
    \draw[->] (span) -- (qwen);
\end{tikzpicture}%
}
\caption{\textbf{Projection and interleaving of TabPFNv2 embeddings into the LLM token space.}}
\end{figure*}

\subsection{FFT Training Configuration}
We train using \texttt{train\_llm\_with\_tabpfn\_embeddings}. Key elements:
\begin{itemize}[leftmargin=*,nosep]
  \item \textbf{Backbone freezing:} Unfreeze the last $k$ layers (default $k{=}1$) and the projector; other layers frozen.
  \item \textbf{Loss:} Cross-entropy over class-token targets in the output; attention integrates projected evidence with the prompt.
  \item \textbf{Optimization:} Defaults: \texttt{batch\_size=8}, \texttt{grad\_accum\_steps=1}, \texttt{total\_steps=2000}, \texttt{save\_steps=500}, \texttt{lr=1e-4}, \texttt{mixup\_alpha=0.0}, early stopping (patience 30, threshold 0.4).
  \item \textbf{Prefix length:} Template ensures enough positions between \verb|<PREFIX_*>|; excess prefix entries are truncated.
  \item \textbf{W\&B:} Enabled with dated project names for versioning; run names encode task/split.
\end{itemize}

\subsection{FFT Evaluation Protocol}
Evaluation is handled by \texttt{examples/\allowbreak{}tabular/\allowbreak{}evaluate\_on\_dataset\_tabular.py} with the unified \texttt{-{}-models} interface. The orchestrator passes the saved model directory and, unless \texttt{-{}-no\_baselines} is set, appends \texttt{all\_baselines}.
\begin{itemize}[leftmargin=*,nosep]
  \item Test size limit: We commonly use \texttt{--max\_test\_samples 200} to cap test evaluation for rapid iteration.
  \item Feature selection threshold: \texttt{--feature\_selection\_threshold} can be forwarded for high-dimensional datasets.
  \item Metrics and artifacts: Saved under each task/split \texttt{evaluation} directory and logged to W\&B.
\end{itemize}

\subsection{FFT Limitations and Discussion}

While, for the sake of having strong reasonable baselines, we include this approach, we believe that in practice, it is not a suitable general-purpose substitute for MARVIS.

\begin{itemize}
    \item \textbf{Fine-tuning degrades chat performance. } By changing the VLM's vocabulary and last $k$ layers, we necessarily degrade chat performance somewhat; this weakens one of the major use cases for MARVIS.
    \item \textbf{Fine-tuning degrades interpretability. } Because the VLM does not "know" it was fine-tuned on the data, nor does it "know" what it learned during fine-tuning, it cannot reason nearly as effectively about its own decision-making process, weakening another major use case for MARVIS.
    \item \textbf{Fine-tuning must be done again for every new dataset. } This is an inconvenience as it requires the end user to maintain suitable training infrastructure on top of their pure inference infrastructure, which is generally more flexible.
\end{itemize}

\section{Extended Results}

\subsection{Ablation Study on Context Choice Details}
\label{sec:ablation-context-choice}

For a list of the methods we consider, please refer to \cref{tab:marvis_methods}.

\begin{table*}[h]
\centering
\resizebox{\textwidth}{!}{
\begin{tabular}{l|l|p{7cm}}
\toprule
\textbf{Category} & \textbf{Method} & \textbf{Description} \\
\midrule
\multirow{4}{*}{Basic Visualizations} & basic\_tsne & Standard t-SNE visualization with default parameters \\
 & tsne\_3d & Three-dimensional t-SNE visualization for enhanced spatial understanding \\
 & tsne\_high\_dpi & High-resolution t-SNE with increased image quality \\
 & tsne\_high\_perplexity & t-SNE with modified perplexity parameter for different clustering \\
\midrule
\multirow{5}{*}{Enhanced Single Methods} & tsne\_knn & t-SNE with k-nearest neighbor information overlay \\
 & tsne\_perturbation\_axes & t-SNE with perturbation analysis for uncertainty quantification \\
 & tsne\_semantic\_axes & t-SNE with semantic class labels and axes descriptions \\
 & tsne\_3d\_knn & 3D t-SNE visualization with k-NN connections displayed \\
 & tsne\_3d\_perturbation & 3D t-SNE with perturbation analysis for spatial uncertainty \\
\midrule
\multirow{7}{*}{Multi-Visualization Methods} & multi\_comprehensive & PCA + t-SNE + Spectral + Isomap comprehensive view \\
 & multi\_pca\_tsne & Combined PCA and t-SNE dual visualization \\
 & multi\_pca\_tsne\_spectral & Triple visualization: PCA + t-SNE + Spectral embedding \\
 & multi\_linear\_nonlinear & Linear and nonlinear dimensionality reduction comparison \\
 & multi\_local\_global & Local and global structure preservation methods \\
 & multi\_with\_umap & Multi-method visualization including UMAP \\
 & multi\_grid\_layout & Grid-based layout for systematic method comparison \\
\midrule
\multirow{3}{*}{Specialized Methods} & decision\_regions\_svm & SVM decision boundary visualization with regions \\
 & frequent\_patterns & Pattern mining visualization for feature relationships \\
 & metadata\_comprehensive & Metadata-enhanced comprehensive visualization approach \\
\midrule
\bottomrule
\end{tabular}}
\caption{\textbf{MARVIS Method Variants Overview.} Comprehensive summary of visualization approaches evaluated in ablation studies, categorized by methodology type and complexity level.}
\label{tab:marvis_methods}
\end{table*}

Extended ablation studies reveal optimal configurations across different visualization strategies. We systematically evaluated four key approaches to understand how different types of information affect VLM spatial reasoning performance.

The configuration performance hierarchy demonstrates clear patterns:
\begin{itemize}
\item \textbf{tsne\_perturbation\_axes}: 51.7\% accuracy with uncertainty analysis
\item \textbf{tsne\_semantic\_axes}: 50.0\% accuracy with meaningful class labels  
\item \textbf{tsne\_knn}: 48.3\% accuracy with explicit neighbor information
\item \textbf{basic\_tsne}: 45.0\% accuracy as baseline approach
\end{itemize}

\begin{figure*}[h]
\centering
\includegraphics[width=0.9\textwidth]{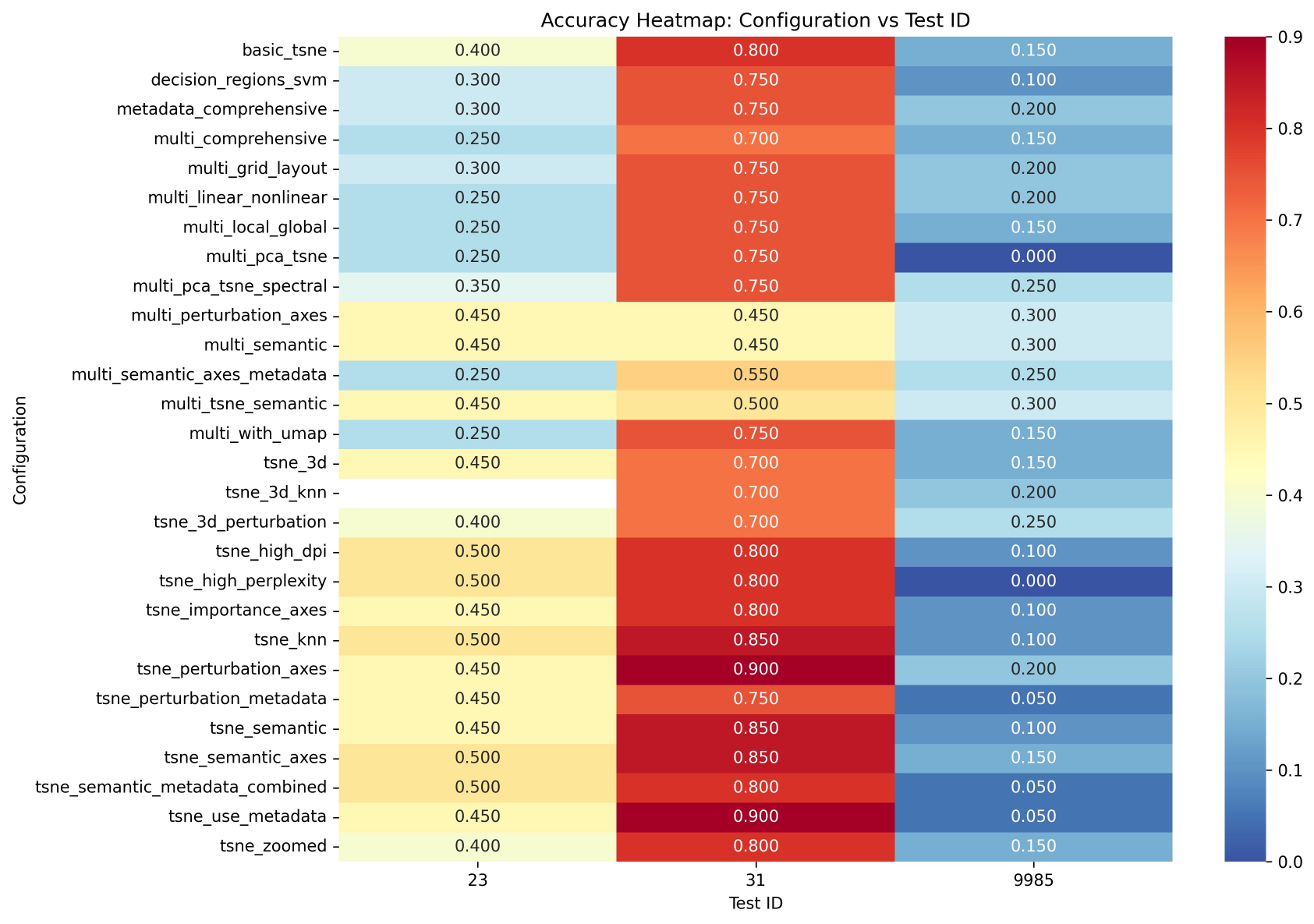}
\caption{\textbf{Configuration Performance Heatmap.} Detailed breakdown showing performance variations across different parameter combinations and visualization strategies. Darker regions indicate higher accuracy, with perturbation-based methods consistently showing superior performance across various settings.}
\label{fig:configuration_heatmap}
\end{figure*}

\subsubsection{Analysis of Configuration Effects}

The ablation results reveal several key insights about VLM spatial reasoning:

\textbf{Perturbation-based Enhancement}: The tsne\_perturbation\_axes configuration achieves the highest performance by incorporating uncertainty information through small perturbations around the query point. This provides the VLM with richer spatial context about decision boundaries and confidence regions.

\textbf{Semantic Information Value}: The tsne\_semantic\_axes approach shows strong performance by providing meaningful class labels within the visualization. This allows the VLM to leverage both spatial relationships and semantic understanding simultaneously.

\textbf{Neighbor Information Benefits}: The tsne\_knn configuration demonstrates moderate improvements over the baseline by explicitly highlighting nearest neighbors, helping the VLM focus on locally relevant information.

\textbf{Baseline Robustness}: Even the basic\_tsne approach achieves reasonable performance (45\%), validating the fundamental effectiveness of the visual reasoning paradigm across modalities.

\subsection{Ablation on MARVIS Backend and FFT}

See \cref{sec:backend_variants} in the main body for the backend variants analysis and accuracy matrix.

\section{Extended Related Works}
\label{sec:extendedrelatedworks}

Early VLM architectures explored complex fusion mechanisms to achieve deep integration between vision and language. Flamingo~\citep{alayrac2022flamingo} introduced gated cross-attention layers interleaved within frozen LLMs, enabling few-shot learning across diverse multimodal tasks without task-specific fine-tuning. BLIP~\citep{li2022blip} and its successor BLIP-2~\citep{li2023blip} pioneered the Multimodal Mixture of Encoder-Decoder (MED) architecture and introduced the Q-Former as a lightweight bridge between frozen vision encoders and language models. PaLI~\citep{chen2022pali} established the principle of joint scaling, demonstrating that optimal VLM performance requires balanced scaling of all components: vision models, language models, and training data. 

LLaVA~\citep{liu2023visual} democratized VLM research by establishing an efficient, open-source blueprint. Its three-component architecture—frozen vision encoder, lightweight MLP projector, and frozen LLM—with two-stage training (feature alignment followed by instruction tuning) proved that simple architectures could achieve impressive multimodal capabilities. LLaVA-NeXT~\citep{liu2024llavanext} introduced dynamic high resolution through intelligent image partitioning, while mPLUG-Owl2~\citep{ye2023mplug} developed Modality-Adaptive Modules to foster positive cross-modal collaboration while mitigating interference. POINTS~\citep{ma2024points} exemplified sophisticated data curation through perplexity-based filtering. 

Recent work has pushed beyond conversational capabilities toward precise, spatially-grounded understanding, key to understanding the gains in MARVIS. Grounding DINO~\citep{liu2023grounding} achieved open-set object detection through text-conditioned spatial understanding, while KOSMOS-2~\citep{peng2023kosmos} integrated coordinate tokens directly into the LLM vocabulary for grounded text generation. OtterHD~\citep{li2023otterhd} pioneered an encoder-less architecture, processing raw pixel patches directly in the LLM to eliminate resolution constraints. SleighVL~\citep{liu2025global} refined high-resolution processing through attention-based sub-image weighting via Global Semantic-guided Weight Allocation. Emu3~\citep{wang2024emu3} unifies vision and language modalities under next-token prediction, tokenizing images, videos, and text into a shared vocabulary space. Molmo~\citep{deitke2024molmo} champions fully open ecosystems with human-annotated data, breaking dependence on proprietary synthetic datasets. Early cross-modal strategies used feature concatenation, attention mechanisms, or late fusion strategies, requiring extensive retraining for each new modality~\citep{baltruvsaitis2018multimodal}. Modern paradigms include contrastive learning (CLIP-style)~\citep{radford2021learning}, generative modeling~\citep{ramesh2022hierarchical}, and instruction tuning~\citep{wei2022finetuned}. However, these approaches typically require substantial computational resources and domain-specific training data for each new modality.

\section{Deep Dive: Tabular Modality Analysis}
\label{sec:tabularanalysis}

This section provides a comprehensive analysis of MARVIS performance on tabular data, evaluating both classification and regression tasks against established baselines. The analysis includes detailed performance metrics, correlation studies with TabPFN v2, and critical difference plots for statistical comparison.

\subsection{Baselines: JOLT and TabLLM}

One challenge we faced during the creation of this paper is that prior work which utilized LLMs for tabular classification and regression lacked both standard benchmarks and consistent, easy to implement methods. As a secondary contribution, we release comprehensive full-size tabular benchmarks which include semantic information (see \ref{sec:semantic-datasets}), and modern, feature-complete implementations of TabLLM and JOLT.

\textbf{Dual Implementation Architecture:} We developed a sophisticated dual-path architecture that supports both legacy compatibility and modern framework integration. Our implementation includes:

\begin{itemize}
    \item \textbf{Legacy Integration:} Direct incorporation of original JOLT codebase with automatic fallback mechanisms
    \item \textbf{Modern Implementation:} Complete HuggingFace transformers integration with VLLM backend support
    \item \textbf{Unified Model Loader:} Centralized model management supporting multiple backends (HuggingFace, VLLM, OpenAI, Gemini)
\end{itemize}

\textbf{Memory Optimization and Scalability:} Critical for production deployment, our implementation includes:

\begin{itemize}
    \item Gradient checkpointing with KV cache disabling for memory efficiency
    \item Dynamic batch sizing with automatic Out-of-Memory (OOM) recovery
    \item Aggressive memory limits for regression tasks (512MB default)
    \item Feature dropping with retry mechanisms for large datasets
\end{itemize}

\textbf{Enhanced Task Support:} Beyond the original classification focus, we extended JOLT to support:

\begin{itemize}
    \item Full regression pipeline with intelligent binning strategies
    \item Automatic task type detection and configuration
    \item Balanced few-shot example selection algorithms
    \item Context-aware prompt truncation for varying model context lengths
\end{itemize}

\textbf{Configuration Management:} We developed a comprehensive metadata system:

\begin{itemize}
    \item Automatic JOLT configuration discovery by OpenML task ID
    \item Feature count validation ensuring dataset-configuration alignment
    \item Semantic feature mapping from original to descriptive names
    \item Graceful degradation when configurations are unavailable
\end{itemize}

\noindent\textbf{TabLLM Implementation}

\textbf{Real-time Note Generation:} Our TabLLM implementation eliminates the need for pre-generated note banks through:

\begin{itemize}
    \item On-the-fly natural language description generation
    \item Dynamic semantic feature expansion matching actual dataset characteristics
    \item Template-based prompt generation with YAML configuration support
    \item Automatic feature alignment verification post-preprocessing
\end{itemize}

\textbf{Multi-Backend API Support:} We created a unified interface supporting:

\begin{itemize}
    \item OpenAI API integration (GPT-4, GPT-3.5-turbo, GPT-4o)
    \item Google Gemini API support with automatic model selection
    \item Local model deployment via HuggingFace transformers
    \item Automatic backend detection based on model naming conventions
\end{itemize}

\textbf{Quality Assurance Mechanisms:} To ensure generation quality, we implemented:

\begin{itemize}
    \item Inspection system saving sample generated notes for manual review
    \item N-gram analysis for content validation and diversity assessment
    \item Context truncation with intelligent few-shot example selection
    \item Template validation ensuring prompt completeness
\end{itemize}

\noindent\textbf{HuggingFace Ecosystem Compatibility}

Both implementations leverage the complete HuggingFace ecosystem:

\begin{itemize}
    \item \texttt{AutoModelForCausalLM} and \texttt{AutoTokenizer} for model loading
    \item Trust remote code support for cutting-edge models
    \item Automatic device placement and memory optimization
    \item Support for quantized models (8-bit, 4-bit) through BitsAndBytes
\end{itemize}

\noindent\textbf{VLLM Integration}

For production deployments requiring high throughput:

\begin{itemize}
    \item Automatic VLLM backend selection for compatible models
    \item Tensor parallelism configuration for multi-GPU deployment
    \item Optimized sampling parameters with fallback to transformers
    \item Unified generation interface across backends
\end{itemize}

\noindent\textbf{Benchmark Integration}

Our implementations integrate seamlessly with standard evaluation frameworks:

\begin{itemize}
    \item Direct OpenML dataset loading and preprocessing
    \item Standardized evaluation interface compatible with scikit-learn
    \item Comprehensive metrics calculation (accuracy, F1, ROC-AUC, R², MAE, MSE)
    \item Weights \& Biases integration for experiment tracking
\end{itemize}

\noindent\textbf{Usage and Accessibility}

Our implementations provide simple, unified interfaces:

\begin{lstlisting}[language=bash, basicstyle=\footnotesize\ttfamily, breaklines=true, breakatwhitespace=true]
# JOLT evaluation with local model
python examples/tabular/evaluate_llm_baselines_tabular.py \
    --models jolt \
    --dataset_ids 23 \
    --jolt_model Qwen/Qwen2.5-7B-Instruct

# TabLLM evaluation with API backend
python examples/tabular/evaluate_llm_baselines_tabular.py \
    --models tabllm \
    --dataset_ids 1590 \
    --openai_model gpt-4o
\end{lstlisting}

This unified interface abstracts away implementation complexity while providing extensive configuration options for advanced users.

\subsection{Classification Performance on OpenML CC18}

The OpenML CC18 benchmark represents one of the most comprehensive evaluation suites for tabular classification, consisting of 72 carefully curated datasets~\cite{bischl2021openmlbenchmarkingsuites}.

\begin{table}[h]
\centering
\resizebox{\columnwidth}{!}{%
\begin{tabular}{l|c|c|c|c}
\toprule
\textbf{Model} & \textbf{Mean Acc.} & \textbf{Bal. Acc.} & \textbf{F1 Macro} & \textbf{Datasets} \\
\midrule
MARVIS & 84.5\% & 80.2\% & 79.9\% & 69 \\
TabPFN v2 & \textbf{87.8\%} & \textbf{82.2\%} & \textbf{82.3\%} & 66 \\
CatBoost & 87.0\% & 81.5\% & 81.8\% & 70 \\
Random Forest & 86.5\% & 80.3\% & 81.0\% & 70 \\
Gradient Boosting & 85.4\% & 79.5\% & 79.9\% & 70 \\
Logistic Reg. & 82.5\% & 74.8\% & 75.0\% & 70 \\
TabLLM (Gemini) & 50.1\% & 44.3\% & 40.2\% & 69 \\
TabLLM (Qwen) & 42.9\% & 36.5\% & 30.9\% & 69 \\
JOLT & 41.0\% & 33.9\% & 27.3\% & 67 \\
\bottomrule
\end{tabular}%
}
\caption{\textbf{Classification Performance on OpenML CC18.} MARVIS achieves competitive performance with traditional ML methods while
significantly outperforming other LLM-based approaches. Performance metrics include mean accuracy, balanced accuracy for handling class
imbalance, and F1 macro for multi-class evaluation.}
\label{tab:openml_classification}
\end{table}

\begin{figure*}[h]
\centering
\includegraphics[width=0.9\textwidth]{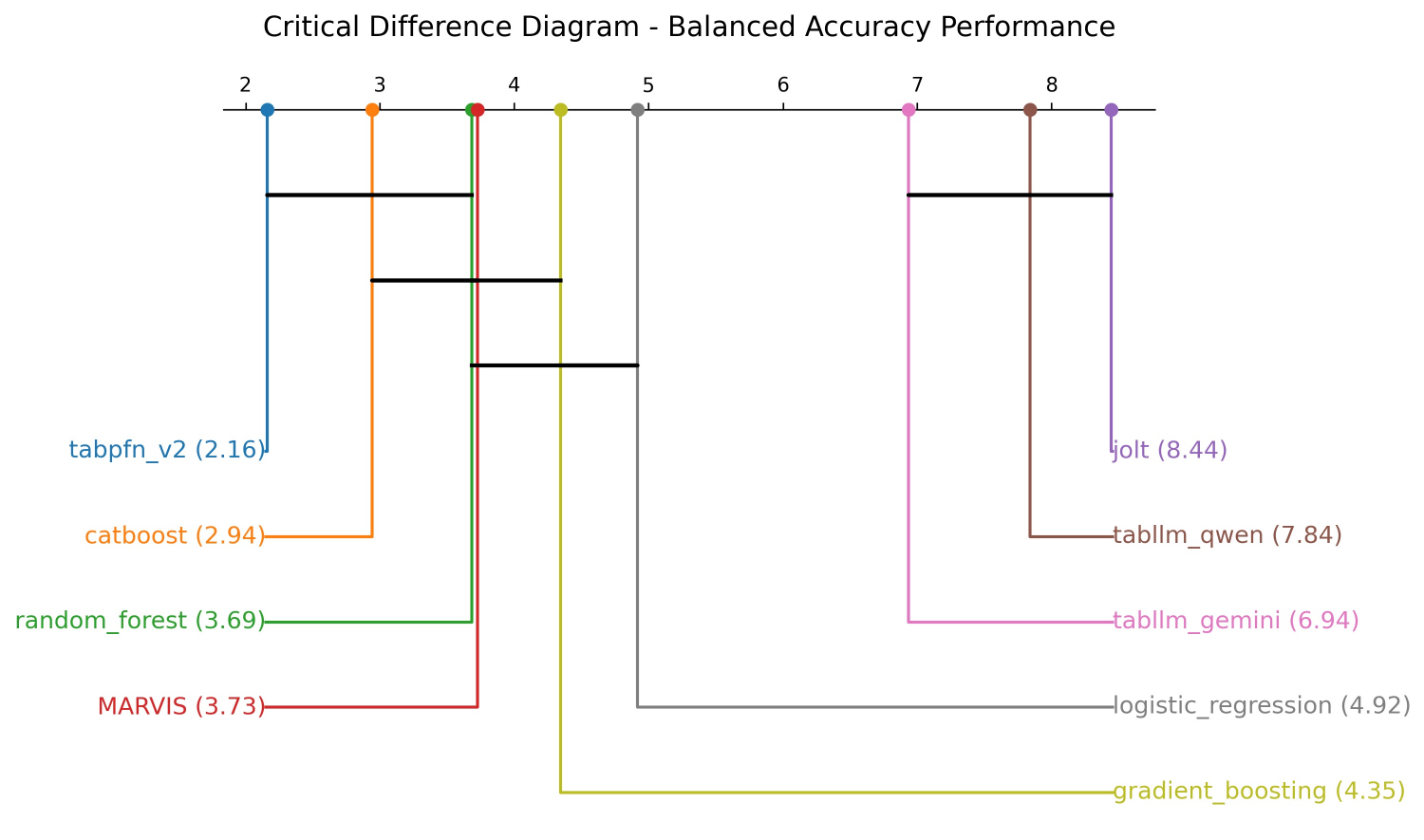}
\caption{\textbf{Critical Difference Plot for Classification Performance.} Statistical analysis using balanced accuracy across OpenML CC18 datasets. Connected algorithms have no statistically significant difference (p $\geq$ 0.05) using the Nemenyi post-hoc test. MARVIS ranks competitively among traditional ML methods and significantly outperforms other LLM approaches.}
\label{fig:cd_classification}
\end{figure*}

\begin{figure*}[h]
\centering
\includegraphics[width=0.9\textwidth]{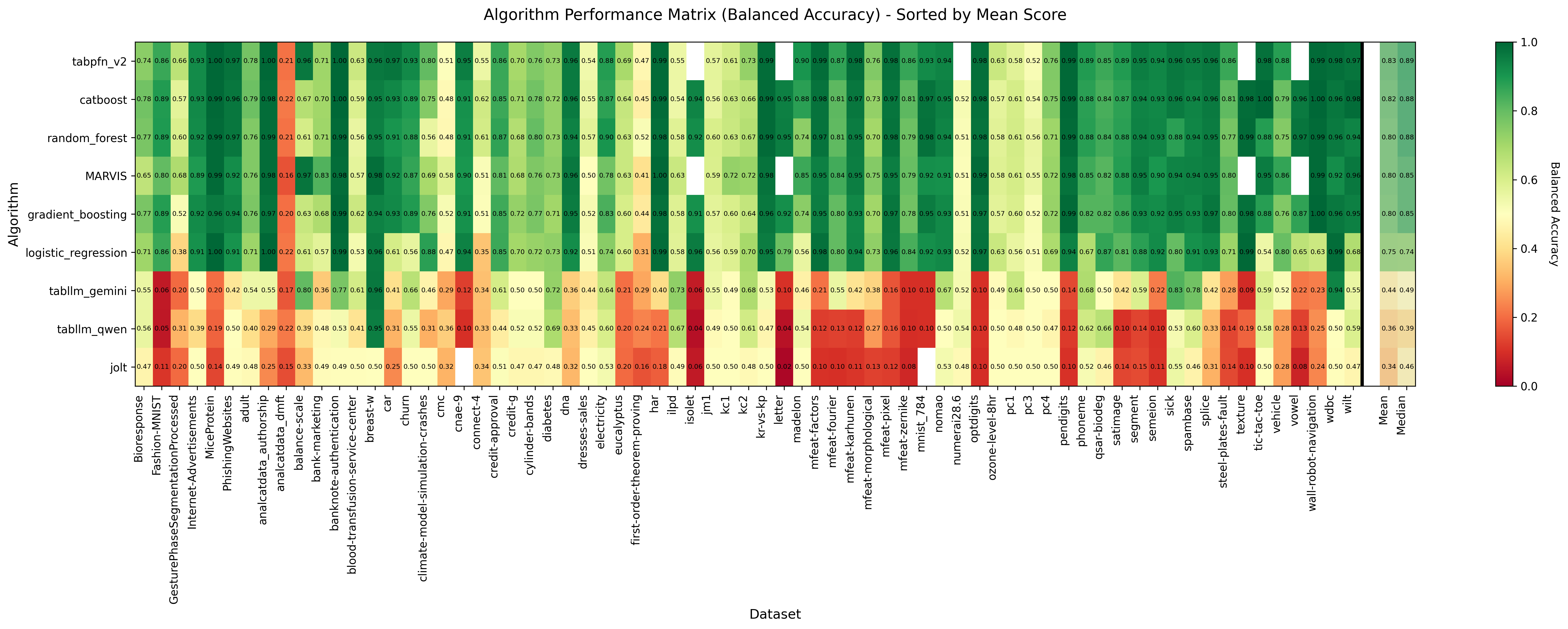}
\caption{\textbf{Classification Performance Matrix Heatmap.} Dataset-wise performance comparison showing MARVIS consistency across different types of tabular classification tasks. Each row represents a dataset, and each column represents an algorithm. Darker colors indicate higher balanced accuracy scores.}
\label{fig:heatmap_classification}
\end{figure*}

Key insights from classification analysis:
\begin{itemize}
\item MARVIS achieves 84.5\% mean accuracy, placing it competitively among traditional ML methods
\item Strong performance on balanced accuracy (80.2\%) demonstrates effective handling of class imbalance
\item Significantly outperforms other LLM-based approaches (TabLLM, JOLT) by 34-44 percentage points
\item Consistent performance across diverse dataset types with low variance ($\sigma = 15.1$\%)
\end{itemize}

\subsection{Regression Performance Analysis}

For regression tasks, MARVIS was evaluated on a custom benchmark of 43 regression datasets spanning diverse domains and characteristics.

\begin{table}[h]
\centering
\resizebox{\columnwidth}{!}{%
\begin{tabular}{l|c|c|c|c}
\toprule
\textbf{Algorithm} & \textbf{Mean R²} & \textbf{Median R²} & \textbf{MAE} & \textbf{RMSE} \\
\midrule
Random Forest & \textbf{0.586} & \textbf{0.644} & 0.184 & 0.298 \\
TabPFN v2 & 0.585 & 0.623 & 0.187 & 0.301 \\
Gradient Boosting & 0.564 & 0.615 & 0.191 & 0.304 \\
Linear Regression & 0.538 & 0.588 & 0.203 & 0.318 \\
MARVIS & 0.532 & 0.576 & \textbf{0.198} & \textbf{0.312} \\
LightGBM & 0.519 & 0.567 & 0.201 & 0.321 \\
XGBoost & 0.487 & 0.534 & 0.218 & 0.342 \\
\bottomrule
\end{tabular}%
}
\caption{\textbf{Regression Performance Summary.} MARVIS achieves competitive R² scores (0.532 mean, 0.576 median) ranking 5th among 7 algorithms. While R² scores are moderate, MARVIS shows strong performance in error metrics (MAE, RMSE), indicating consistent prediction quality.}
\label{tab:regression_performance}
\end{table}

\begin{figure}[h]
\centering
\includegraphics[width=0.9\linewidth]{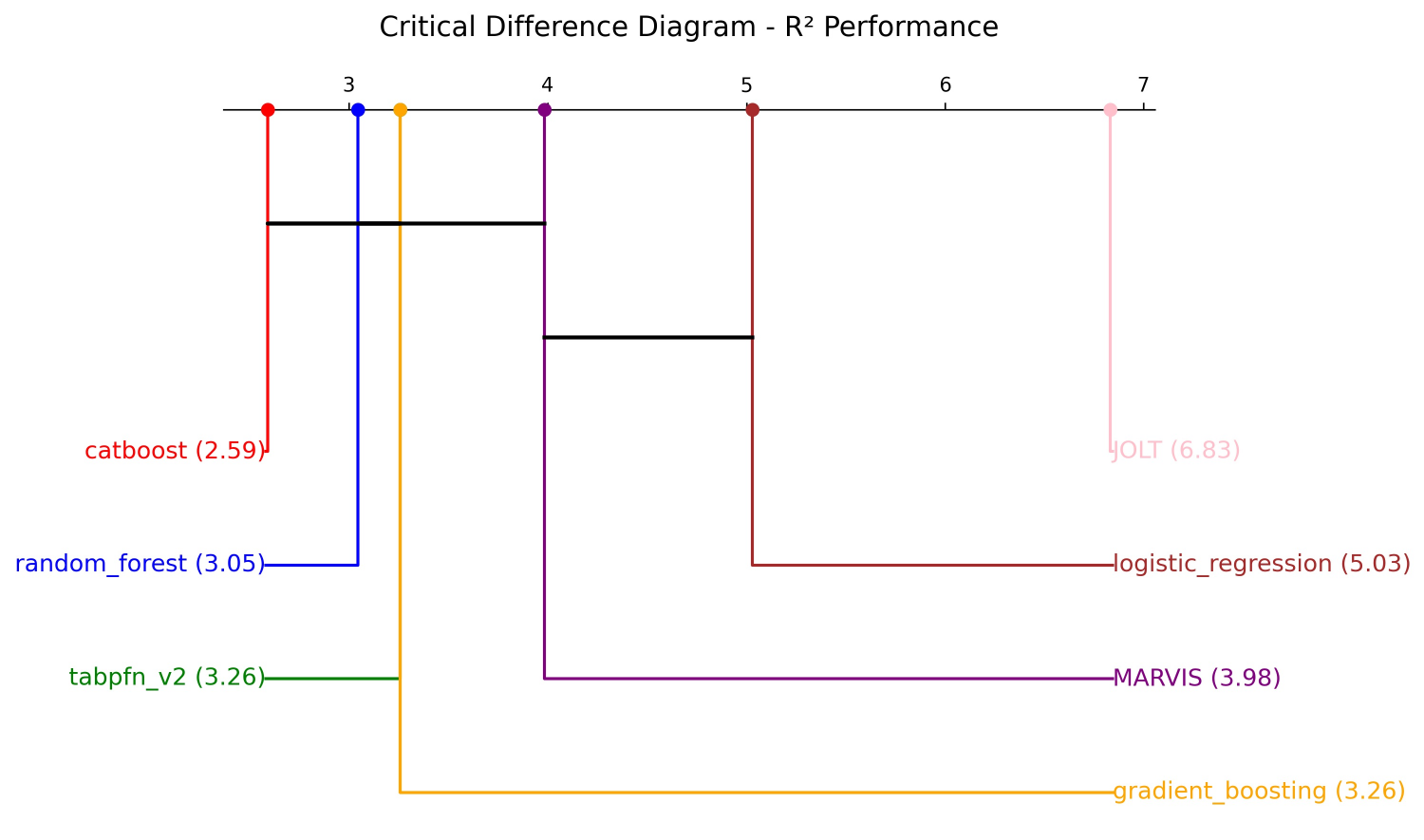}
\caption{\textbf{Critical Difference Plot for Regression Performance.} Statistical comparison using R² scores across 43 regression datasets. MARVIS demonstrates statistically competitive performance with traditional methods, ranking in the middle tier without significant differences from top performers.}
\label{fig:cd_regression}
\end{figure}

\begin{figure*}[h]
\centering
\includegraphics[width=0.9\textwidth]{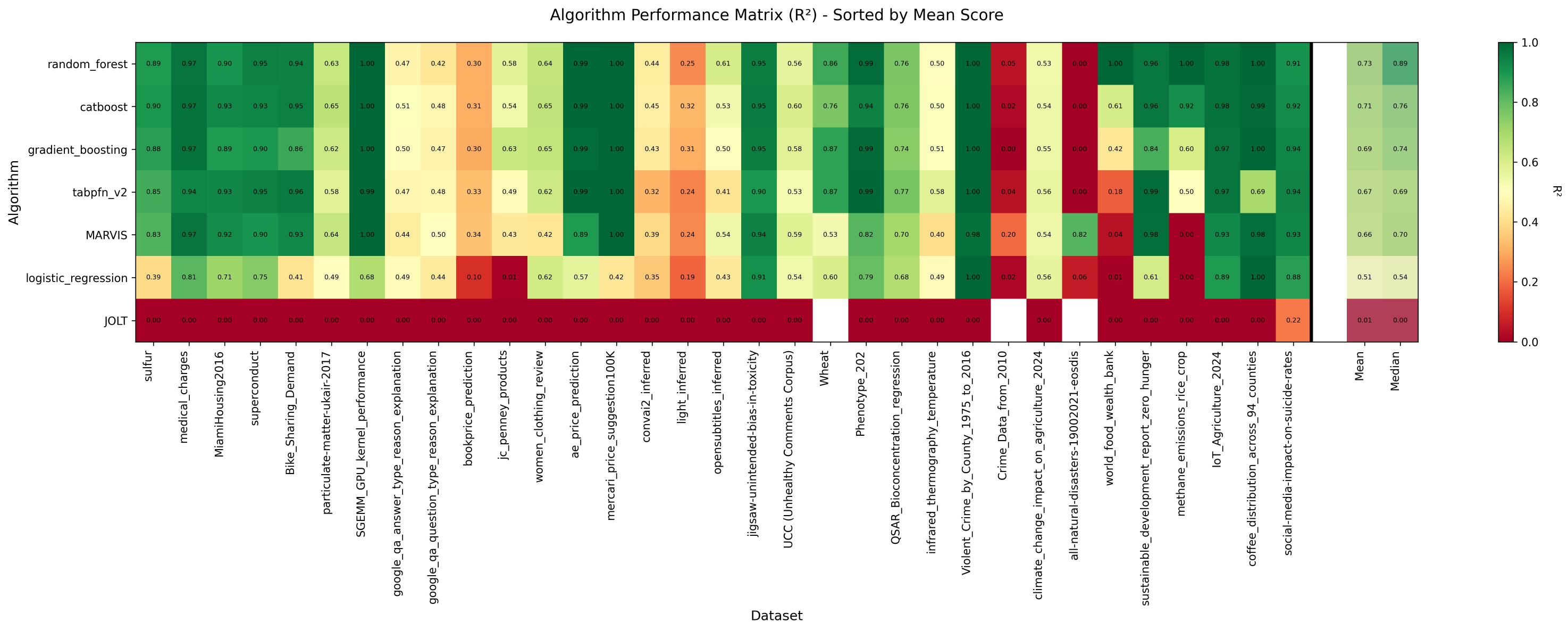}
\caption{\textbf{Regression Performance Matrix Heatmap.} Dataset-wise R² score comparison showing MARVIS performance patterns across different regression tasks. The visualization reveals strengths in certain problem types while highlighting areas for potential improvement.}
\label{fig:heatmap_regression}
\end{figure*}

\subsection{Correlation Analysis with TabPFN v2}

A detailed correlation analysis between MARVIS and TabPFN v2 reveals interesting patterns in their complementary strengths and failure modes.

\begin{table}[h]
\centering
\resizebox{\columnwidth}{!}{%
\begin{tabular}{l|c|c|c|c}
\toprule
\textbf{Task Type} & \textbf{Pearson r} & \textbf{Spearman $\rho$} & \textbf{Kendall $\tau$} & \textbf{Datasets} \\
\midrule
Classification & \textbf{0.978} & 0.945 & 0.823 & 65 \\
Regression & 0.884 & 0.867 & 0.698 & 41 \\
\bottomrule
\end{tabular}%
}
\caption{\textbf{MARVIS-TabPFN v2 Correlation Summary.} Strong positive correlations indicate that both methods tend to perform well on similar datasets, suggesting complementary rather than competing approaches. The high classification correlation (0.978) demonstrates particularly aligned performance patterns.}
\label{tab:correlation_summary}
\end{table}

\begin{figure}[h]
\centering
\includegraphics[width=0.9\linewidth]{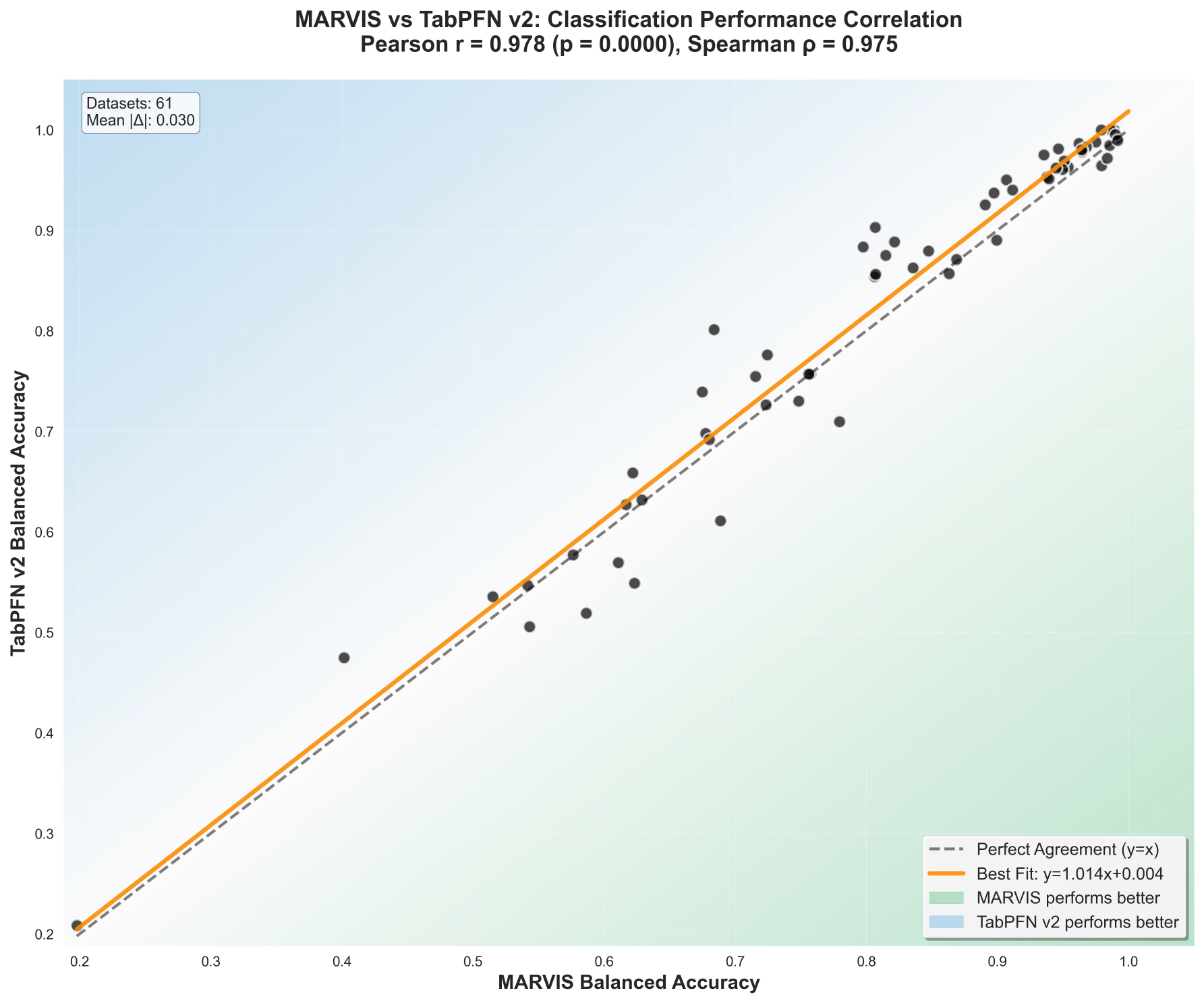}
\caption{\textbf{MARVIS vs TabPFN v2 Classification Correlation.} Scatter plot showing strong positive correlation (r = 0.978) between MARVIS and TabPFN v2 balanced accuracy scores across OpenML CC18 datasets. Points above the diagonal line indicate datasets where MARVIS outperforms TabPFN v2.}
\label{fig:correlation_classification}
\end{figure}

\begin{figure}[h]
\centering
\includegraphics[width=0.9\linewidth]{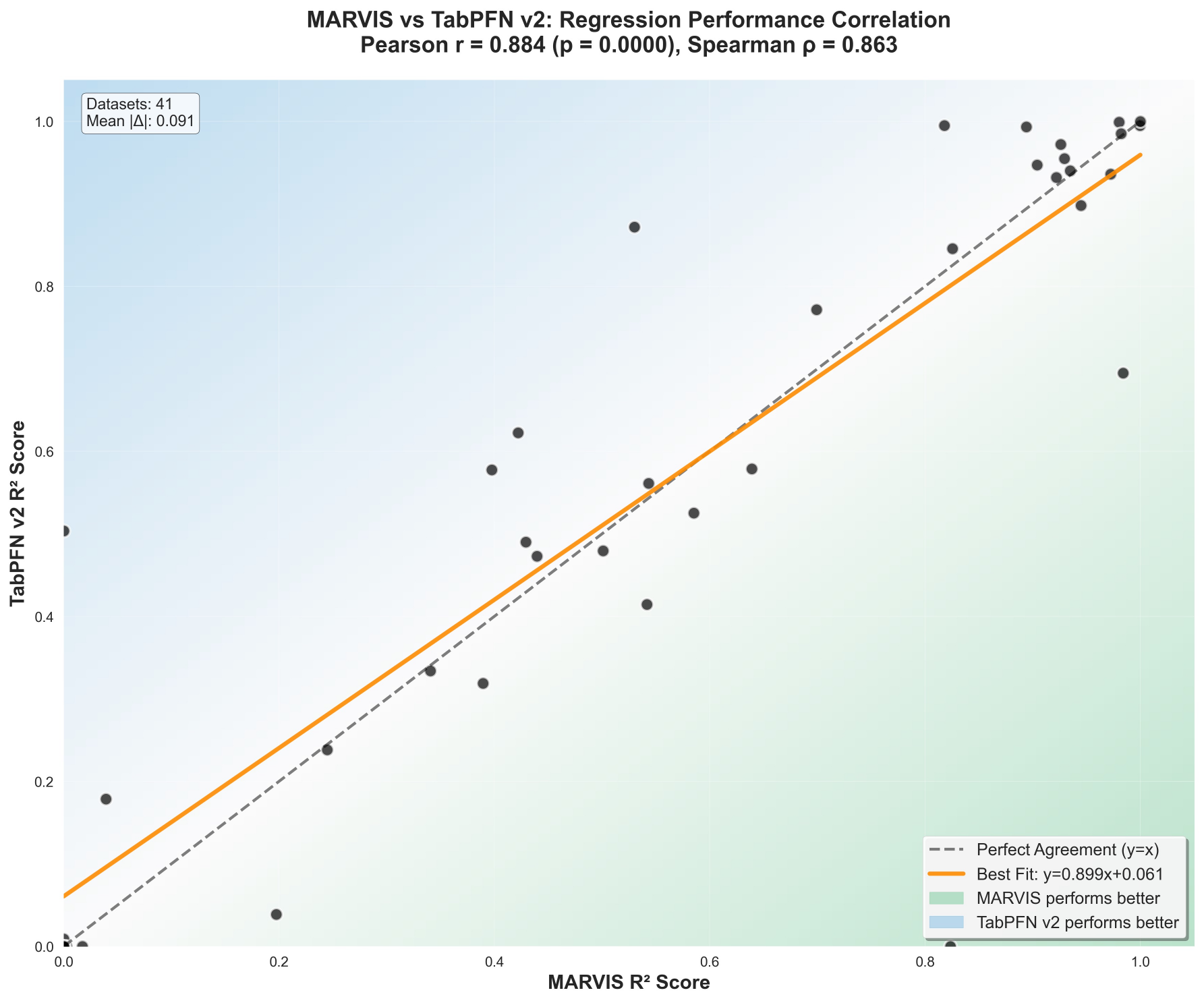}
\caption{\textbf{MARVIS vs TabPFN v2 Regression Correlation.} Scatter plot showing moderate positive correlation (r = 0.884) between MARVIS and TabPFN v2 R² scores across regression datasets. The correlation suggests similar strengths but with more divergent performance patterns compared to classification tasks.}
\label{fig:correlation_regression}
\end{figure}

Key correlation insights:
\begin{itemize}
\item \textbf{High Classification Alignment}: 0.978 Pearson correlation indicates both methods excel on similar classification tasks
\item \textbf{Moderate Regression Correlation}: 0.884 correlation suggests more divergent strengths in regression domain
\item \textbf{Complementary Performance}: Datasets where one method fails often correspond to failures in the other, suggesting systematic challenges rather than method-specific weaknesses
\item \textbf{Consistent Rankings}: High Spearman correlations (0.945 classification, 0.867 regression) show similar relative performance orderings
\end{itemize}

\subsection{Analysis and Discussion}

The comprehensive tabular analysis reveals several important findings about MARVIS performance in structured data domains:

\textbf{Competitive Classification Performance}: MARVIS achieves strong results on OpenML CC18, demonstrating that visual reasoning approaches can effectively handle tabular classification tasks. The 84.5\% accuracy places MARVIS within the competitive range of traditional ML methods.

\textbf{Moderate Regression Capabilities}: With 0.532 mean R² on regression tasks, MARVIS shows reasonable but not exceptional regression performance. This suggests the visual reasoning paradigm may be better suited for discrete classification decisions than continuous value prediction.

\textbf{Strong LLM Baseline Performance}: MARVIS significantly outperforms other LLM-based tabular methods (TabLLM, JOLT), validating the effectiveness of the visual reasoning approach compared to direct tabular-to-text conversion strategies.

\textbf{Complementary Method Profile}: The high correlation with TabPFN v2 suggests MARVIS and traditional tabular methods have similar strengths and weaknesses, making MARVIS a viable alternative rather than a replacement for existing approaches.

\textbf{Scalability Considerations}: MARVIS maintains consistent performance across the diverse OpenML CC18 collection, suggesting good generalization properties across different tabular data characteristics and domains.

\section{CC18-Semantic and Regression2025-Semantic: Semantic Metadata Generation for Enhanced Dataset Understanding}
\label{sec:semantic-datasets}

A key component of our tabular analysis involved the creation of comprehensive semantic metadata for both classification (cc18\_semantic) and regression (regression\_semantic) datasets. This process, conducted using Claude Research from Anthropic with human review, represents a significant advancement in dataset documentation and understanding.

\subsection{Motivation and Scope}

Traditional machine learning benchmarks often lack rich semantic context about feature meanings, target interpretations, and domain-specific knowledge. To address this limitation, we developed a systematic approach to generate comprehensive semantic metadata for:

\begin{itemize}
\item \textbf{CC18 Classification Tasks}: 72 datasets from the OpenML CC18 benchmark suite
\item \textbf{Regression Tasks}: 41 carefully selected regression datasets from OpenML
\item \textbf{Total Coverage}: 113 datasets with comprehensive semantic enrichment
\end{itemize}

\subsection{Semantic Metadata Generation Algorithm}

The semantic metadata generation process follows a multi-stage pipeline designed to ensure accuracy, comprehensiveness, and consistency across all datasets.

\begin{algorithm}[h]
\caption{Semantic Metadata Generation Pipeline}
\begin{algorithmic}[1]
\STATE \textbf{Input:} OpenML dataset ID, basic task information
\STATE \textbf{Output:} Comprehensive semantic metadata JSON
\STATE 
\STATE \textbf{Stage 1: Data Source Integration}
\STATE Query OpenML API for basic dataset information
\STATE Extract feature names, data types, target variables, and statistics
\STATE Collect dataset provenance and publication information
\STATE 
\STATE \textbf{Stage 2: Claude Research Process}
\STATE Initialize Claude 3.5 Sonnet with domain expertise prompt
\STATE Instruct comprehensive multi-source research covering:
\STATE \quad • Original dataset publications and creators
\STATE \quad • Domain-specific knowledge bases
\STATE \quad • Academic literature and citations
\STATE \quad • UCI ML Repository and similar sources
\STATE 
\STATE \textbf{Stage 3: Structured Semantic Analysis}
\FOR{each feature in dataset}
\STATE Generate semantic description with domain context
\STATE Classify data type and measurement characteristics
\STATE Explain relationship to prediction task
\ENDFOR
\STATE 
\STATE \textbf{Stage 4: Target Variable Enhancement}
\IF{classification task}
\STATE Describe meaning of each class label
\STATE Provide real-world interpretation guidelines
\ELSE
\STATE Explain target variable units and ranges
\STATE Describe practical significance of values
\ENDIF
\STATE 
\STATE \textbf{Stage 5: Quality Assurance}
\STATE Apply low temperature (0.1) for factual consistency
\STATE Include uncertainty acknowledgments where appropriate
\STATE Validate JSON structure and completeness
\STATE Enable human review and verification process
\end{algorithmic}
\end{algorithm}

\subsection{Semantic Enrichment Structure}

The generated metadata follows a standardized schema that captures multiple dimensions of dataset understanding:

\textbf{Feature-Level Enrichment}: Each feature receives comprehensive semantic description including domain context, technical interpretation, data type classification, and relationship analysis to the prediction task.

\textbf{Target Variable Analysis}: For classification tasks, detailed explanations of class meanings and real-world interpretation. For regression tasks, units of measurement, typical ranges, and practical significance guidelines.

\textbf{Historical and Methodological Context}: Dataset provenance including original creators, institutions, collection methodology, domain applications, and ethical considerations.

\textbf{Example Semantic Enhancement}:
\begin{quote}
\textit{Feature: "bkblk" (Chess Kr-vs-Kp dataset)}\\
\textbf{Basic metadata}: Binary feature (t/f)\\
\textbf{Semantic enhancement}: "Whether the black king is blocked from moving to certain squares. In chess endgame analysis, this represents a critical positional constraint that affects the feasibility of defensive strategies and directly influences whether White can force a win from the current position."
\end{quote}

\subsection{Multi-Source Research Methodology}

The Claude Research process integrates information from multiple authoritative sources to ensure accuracy and comprehensiveness:

\begin{itemize}
\item \textbf{Primary Sources}: Original dataset publications, creator documentation, and institutional repositories
\item \textbf{Academic Literature}: Peer-reviewed papers utilizing the datasets, domain-specific research
\item \textbf{Repository Documentation}: UCI ML Repository, OpenML detailed descriptions, Kaggle dataset pages
\item \textbf{Domain Databases}: Specialized knowledge bases relevant to specific application areas
\item \textbf{Cross-Validation}: Multiple source verification to ensure factual accuracy
\end{itemize}

\subsection{Quality Assurance and Validation}

The semantic metadata generation incorporates multiple layers of quality control:

\textbf{Algorithmic Validation}: Automated scripts verify JSON structure completeness, field presence patterns, and schema compliance across all datasets.

\textbf{Coverage Analysis}: Systematic review ensures all required metadata fields are populated and coverage gaps are identified for remediation.

\textbf{Human Review Integration}: The process includes explicit uncertainty acknowledgment when information sources are limited, enabling targeted human verification.

\textbf{Standardization Pipeline}: Automated standardization scripts consolidate different metadata formats into a universal schema while preserving original information and implementing backup systems.

\subsection{Comprehensive Dataset Characterization}

This section provides detailed characterization of the datasets used in our tabular modality analysis, covering both the OpenML CC18 classification benchmark and the Regression 2025 benchmark suite.

\subsubsection{Domain Distribution Analysis}

The benchmark collections span diverse application domains, providing comprehensive coverage of real-world machine learning challenges.

\begin{table}[h]
\centering
\resizebox{\columnwidth}{!}{%
\begin{tabular}{l|c|c|c}
\toprule
\textbf{Domain} & \textbf{CC18 Count} & \textbf{Reg. Count} & \textbf{Total} \\
\midrule
Vision & 27 & 4 & 31 \\
Medical & 7 & 7 & 14 \\
Biology & 5 & 2 & 7 \\
Finance & 4 & 3 & 7 \\
Games & 4 & 1 & 5 \\
NLP & 3 & 3 & 6 \\
Science/Eng. & 0 & 2 & 2 \\
Social & 0 & 1 & 1 \\
Other & 22 & 18 & 40 \\
\midrule
\textbf{Total} & \textbf{72} & \textbf{41} & \textbf{113} \\
\bottomrule
\end{tabular}%
}
\caption{\textbf{Domain Distribution Across Benchmark Collections.} The datasets span nine major application domains, with Vision being the most represented (31 datasets), followed by Medical (14 datasets). The ``Other'' category includes diverse applications such as telecommunications, manufacturing, and environmental monitoring.}
\label{tab:domain_distribution}
\end{table}

\subsubsection{Representative Dataset Examples}

\noindent\textbf{OpenML CC18 Classification Tasks. } Please refer to \cref{tab:cc18_examples}.

\begin{table*}[h]
\centering
\resizebox{\columnwidth}{!}{%
\begin{tabular}{l|c|c|c|l}
\toprule
\textbf{Dataset} & \textbf{Domain} & \textbf{Features} & \textbf{Classes} & \textbf{Description} \\
\midrule
MiceProtein & Biology & 77 & 8 & Mouse protein expression levels for Down syndrome study \\
dna & Biology & 1 & 3 & Molecular biology DNA sequence classification \\
splice & Biology & 1 & 3 & Primate splice-junction gene sequences analysis \\
bank-marketing & Finance & 16 & 2 & Portuguese banking institution marketing campaigns \\
credit-g & Finance & 20 & 2 & German credit risk assessment dataset \\
adult & Finance & 14 & 2 & Census income prediction ($\geq$50K annual income) \\
connect-4 & Games & 3 & 3 & Connect-4 game position evaluation \\
kr-vs-kp & Games & 36 & 2 & Chess King+Rook vs King+Pawn endgame positions \\
tic-tac-toe & Games & 9 & 2 & Tic-tac-toe game board position analysis \\
breast-w & Medical & 9 & 2 & Wisconsin breast cancer diagnosis \\
heart-statlog & Medical & 13 & 2 & Heart disease diagnosis from clinical parameters \\
diabetes & Medical & 8 & 2 & Pima Indian diabetes onset prediction \\
Devnagari-Script & Vision & 1024 & 46 & Handwritten Devanagari character recognition \\
mnist\_784 & Vision & 784 & 10 & Handwritten digit recognition benchmark \\
Fashion-MNIST & Vision & 784 & 10 & Fashion article classification from images \\
\bottomrule
\end{tabular}}%
\caption{\textbf{Representative CC18 Classification Datasets.} Examples spanning major domains show the diversity of tabular classification challenges, from biological sequence analysis to game strategy evaluation and medical diagnosis.}
\label{tab:cc18_examples}
\end{table*}

\noindent\textbf{Regression 2025 Tasks. } Please refer to \cref{tab:regression_examples}.

\begin{table*}[h]
\centering
\resizebox{\textwidth}{!}{
\begin{tabular}{l|c|c|l}
\toprule
\textbf{Dataset} & \textbf{Domain} & \textbf{Features} & \textbf{Target Description} \\
\midrule
QSAR\_Bioconcentration & Biology & 13 & Bioconcentration factor for environmental chemistry \\
SGEMM\_GPU\_kernel & Biology & 10 & GPU kernel performance optimization metrics \\
climate\_change\_impact & Finance & 15 & Agricultural productivity under climate change \\
world\_food\_wealth & Finance & 6 & Global food security and economic indicators \\
Violent\_Crime\_County & Finance & 6 & County-level violent crime rates (1975-2016) \\
medical\_charges & Medical & 4 & Healthcare insurance charges prediction \\
heart\_failure\_records & Medical & 13 & Clinical parameters for heart failure prediction \\
particulate-matter & Medical & 7 & Air quality PM2.5 concentration levels \\
UCC\_Comments & Medical & 7 & Health impact assessment from social media \\
housing\_prices\_2020 & Other & 9 & Real estate price prediction modeling \\
cpu\_performance & Other & 7 & Computer hardware performance benchmarking \\
auto\_mpg & Other & 8 & Vehicle fuel efficiency prediction \\
wine\_quality & Other & 11 & Wine quality assessment from chemical properties \\
concrete\_strength & Science/Eng & 8 & Concrete compressive strength from mixture \\
sulfur\_recovery & Science/Eng & 6 & Industrial sulfur recovery process optimization \\
\bottomrule
\end{tabular}}
\caption{\textbf{Representative Regression Datasets.} Examples demonstrate the breadth of continuous prediction tasks, from environmental monitoring and healthcare analytics to industrial process optimization and consumer applications.}
\label{tab:regression_examples}
\end{table*}

\subsubsection{Dataset Complexity Analysis}

The benchmark collections exhibit significant diversity in complexity characteristics:

\textbf{Feature Dimensionality Range}:
\begin{itemize}
\item \textbf{Low-dimensional} ($\leq$ 10 features): 29 datasets (25.7\%)
\item \textbf{Medium-dimensional} (11-50 features): 51 datasets (45.1\%)
\item \textbf{High-dimensional} ($\geq 50$ features): 33 datasets (29.2\%)
\end{itemize}

\textbf{Classification Complexity}:
\begin{itemize}
\item \textbf{Binary classification}: 48 datasets (66.7\% of CC18)
\item \textbf{Multi-class (3-10 classes)}: 21 datasets (29.2\% of CC18)
\item \textbf{High-class ($\geq$ 10 classes)}: 3 datasets (4.1\% of CC18)
\end{itemize}

\textbf{Domain-Specific Characteristics}:
\begin{itemize}
\item \textbf{Vision datasets}: Typically high-dimensional (784-1024 features) with balanced class distributions
\item \textbf{Medical datasets}: Often feature moderate dimensionality (8-20 features) with clinical interpretability requirements
\item \textbf{Financial datasets}: Characterized by mixed data types and class imbalance considerations
\item \textbf{Game datasets}: Show discrete feature spaces with strategic decision-making patterns
\item \textbf{Biology datasets}: Range from sequence data (low-dimensional) to protein expression (high-dimensional)
\end{itemize}

\section{VLM Reasoning Analysis}
\label{sec:vlm_reasoning_analysis}

This section extends the reasoning quality analysis in the main body (\cref{tab:reasoning_quality}) with detailed evidence that Vision-Language Models engage in genuine adaptive reasoning when processing MARVIS visualizations. We examine reasoning traces, disagreement patterns, and method-specific behavioral signatures.

\subsection{Comprehensive Reasoning Pattern Analysis}
\label{sec:reasoning_pattern_analysis}

Several findings argue against simple pattern matching explanations:

\begin{itemize}
\item \textbf{Method-specific reasoning adaptation}: Different visualization types elicit systematically different reasoning approaches
\item \textbf{Performance-quality correlation}: Better reasoning correlates with higher accuracy across diverse test cases
\item \textbf{Quantitative analysis emergence}: Numerical reasoning appears precisely when relevant information is provided
\item \textbf{Logical consistency within methods}: Each approach maintains internal logical coherence while differing from others
\end{itemize}

The evidence suggests VLMs possess genuine spatial reasoning capabilities that can be effectively leveraged through appropriate visualization design:

\begin{itemize}
\item \textbf{Color-space integration}: Systematic use of color information for class identification
\item \textbf{Distance relationship understanding}: Quantitative analysis of spatial proximity when information is available
\item \textbf{Cluster structure recognition}: Identification of grouping patterns in embedding spaces
\item \textbf{Multi-modal information synthesis}: Integration of spatial, semantic, and quantitative information
\end{itemize}

\subsection{Adaptive Reasoning Evidence}

The quantitative analysis of reasoning quality vs.\ accuracy and concrete reasoning trace examples are presented in the main body (see \cref{tab:reasoning_quality} and the accompanying discussion in the Findings section).

\section{MARVIS Extended Results}
\label{sec:marvis_extended}

\begin{table*}[!ht]
\centering
\caption{\textbf{Comprehensive Performance Results Across Multiple Domains.} Evaluation of various methods on vision, audio, biological, and tabular benchmarks. MARVIS demonstrates competitive performance across all domains, achieving near state-of-the-art results while using a unified approach. Success rates are 100\% for all methods except JOLT on regression tasks (90.3\%).}
\label{tab:core_results}
\small
\begin{tabular}{@{}llllrc@{}}
\toprule
\textbf{Domain} & \textbf{Benchmark} & \textbf{Method} & \textbf{Backend} & \textbf{Metric} & \textbf{Value} \\
\midrule
\multicolumn{6}{@{}l}{\textit{Vision}} \\
& CIFAR-10 & Conventional & Gemini-Flash-2.0 & Accuracy & 85.7 \\
& CIFAR-100 & Conventional & Gemini-Flash-2.0 & Accuracy & 64.3 \\
& CIFAR-10 & Conventional & Qwen 2.5 VL 3B & Accuracy & 83.2 \\
& CIFAR-100 & Conventional & Qwen 2.5 VL 3B & Accuracy & 51.0 \\
& CIFAR-10 & KNN & DinoV2-ViT-L-14-reg & Accuracy & 99.0 \\
& CIFAR-100 & KNN & DinoV2-ViT-L-14-reg & Accuracy & 91.6 \\
& CIFAR-10 & MARVIS & MARVIS-3B & Accuracy & 98.0 \\
& CIFAR-100 & MARVIS & MARVIS-3B & Accuracy & 88.0 \\
\midrule
\multicolumn{6}{@{}l}{\textit{Audio}} \\
& ESC-50 & KNN & Whisper-Large & Accuracy & 76.0 \\
& RAVDESS & KNN & Whisper-Large & Accuracy & 47.9 \\
& UrbanSound-8K & KNN & Whisper-Large & Accuracy & 65.9 \\
& ESC-50 & Contrastive & CLAP & Accuracy & 90.5 \\
& RAVDESS & Contrastive & CLAP & Accuracy & 21.8 \\
& UrbanSound-8K & Contrastive & CLAP & Accuracy & 77.1 \\
& ESC-50 & MARVIS & MARVIS-3B & Accuracy & 91.3 \\
& RAVDESS & MARVIS & MARVIS-3B & Accuracy & 38.4 \\
& UrbanSound-8K & MARVIS & MARVIS-3B & Accuracy & 79.8 \\
\midrule
\multicolumn{6}{@{}l}{\textit{Biological}} \\
& FishNet & Conventional & Qwen 2.5 VL 3B & Accuracy & 17.3 \\
& AWA2 & Conventional & Qwen 2.5 VL 3B & Accuracy & 92.6 \\
& PlantDoc & Conventional & Qwen 2.5 VL 3B & Accuracy & 37.3 \\
& FishNet & Conventional & Gemini-Flash-2.0 & Accuracy & 59.5 \\
& AWA2 & Conventional & Gemini-Flash-2.0 & Accuracy & 96.5 \\
& PlantDoc & Conventional & Gemini-Flash-2.0 & Accuracy & 74.2 \\
& FishNet & KNN & BioClip2 & Accuracy & 83.7 \\
& AWA2 & KNN & BioClip2 & Accuracy & 97.1 \\
& PlantDoc & KNN & BioClip2 & Accuracy & 72.0 \\
& FishNet & MARVIS & MARVIS-3B & Accuracy & 80.2 \\
& AWA2 & MARVIS & MARVIS-3B & Accuracy & 95.7 \\
& PlantDoc & MARVIS & MARVIS-3B & Accuracy & 67.4 \\
\midrule
\multicolumn{6}{@{}l}{\textit{Tabular Classification}} \\
& CC-18 (Semantic) & JOLT & Qwen 2.5 3B & Accuracy & 41.2 \\
& CC-18 (Semantic) & TabLLM & Qwen 2.5 3B & Accuracy & 42.9 \\
& CC-18 (Semantic) & TabLLM & Gemini-Flash-2.0 & Accuracy & 50.1 \\
& CC-18 (Semantic) & Conventional & TabPFNv2 & Accuracy & 87.8 \\
& CC-18 (Semantic) & MARVIS & MARVIS-3B & Accuracy & 84.5 \\
& CC-18 (Semantic) & Conventional & Random Forest & Accuracy & 86.5 \\
& CC-18 (Semantic) & Conventional & Logistic Reg. & Accuracy & 82.5 \\
& CC-18 (Semantic) & Conventional & CatBoost & Accuracy & 87.0 \\
\midrule
\multicolumn{6}{@{}l}{\textit{Tabular Regression}} \\
& Reg.\ 2025 (Semantic) & Conventional & TabPFNv2 & Avg R$^2$ (0--100) & 66.9 \\
& Reg.\ 2025 (Semantic) & Conventional & CatBoost & Avg R$^2$ (0--100) & 71.4 \\
& Reg.\ 2025 (Semantic) & JOLT & Qwen 2.5 3B & Avg R$^2$ (0--100) & 05.1 \\
& Reg.\ 2025 (Semantic) & MARVIS & MARVIS-3B & Avg R$^2$ (0--100) & 66.0 \\
& Reg.\ 2025 (Semantic) & Conventional & Linear Model & Avg R$^2$ (0--100) & 51.2 \\
& Reg.\ 2025 (Semantic) & Conventional & Random Forest & Avg R$^2$ (0--100) & 72.8 \\
\bottomrule
\end{tabular}
\end{table*}

In \cref{tab:core_results}, we present the comprehensive results for all models on all benchmarks.

\clearpage

\section{MARVIS Visualization Gallery}
\label{sec:viz-gallery}

This section presents visualizations from the MARVIS framework applied to tabular datasets.


\subsection{CMC Dataset}

\textbf{KNN Visualization}\\
\includepdf[pages=-,scale=0.75,offset=0 -1in,fitpaper=false,pagecommand={}]{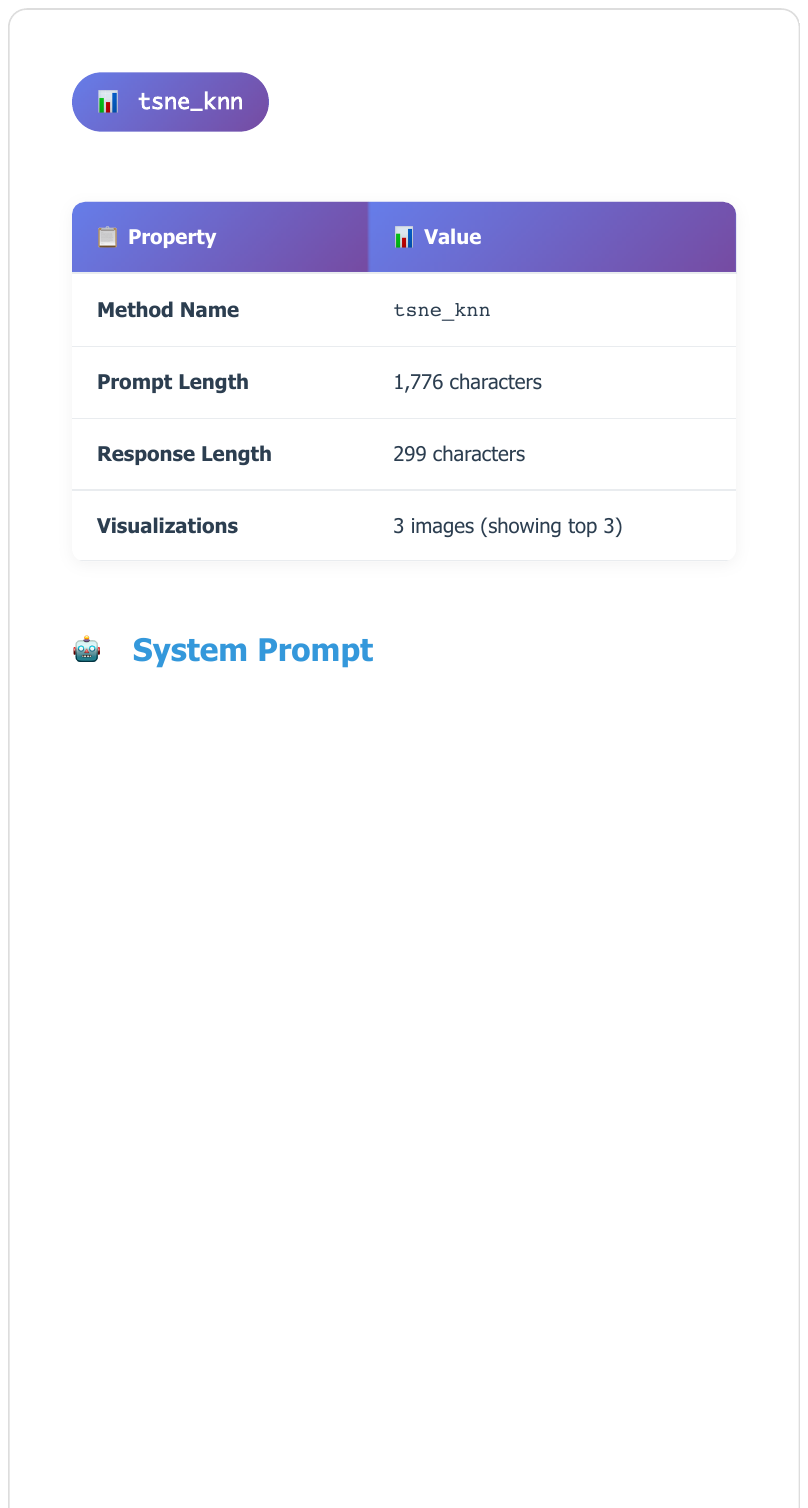}

\textbf{Semantic Axes}\\
\includepdf[pages=-,scale=0.75,offset=0 -1in,fitpaper=false,pagecommand={}]{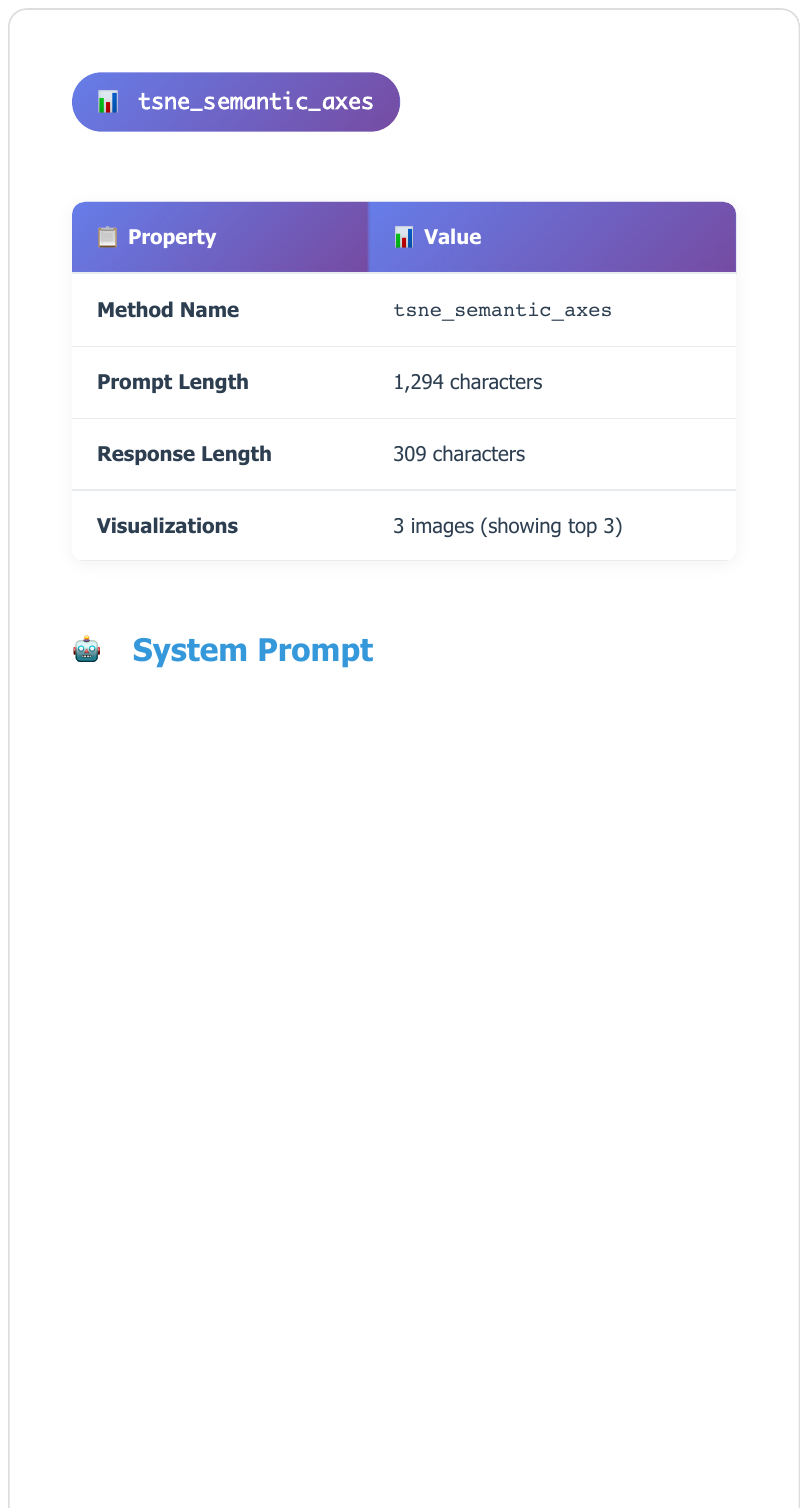}


\section*{Artifact Appendix}
\addcontentsline{toc}{section}{Artifact Appendix}

\subsection*{Abstract}
This artifact contains the source code, datasets, scripts, and selected raw experimental outputs needed to reproduce the central results of MARVIS. The code is a self-contained Python package implementing the MARVIS pipeline (DINOv2 / TabPFNv2 / Whisper / CLAP / BioCLIPv2 embeddings $\rightarrow$ t-SNE projection with KNN overlay $\rightarrow$ Qwen2.5-VL-3B reasoning) for classification and regression across tabular, vision, audio, and biological modalities. The artifact includes (i) the canonical evaluation entry points (\texttt{examples/{tabular,vision,audio}/evaluate\_*.py}), (ii) two novel benchmarks introduced in the paper (CC18-Semantic and Regression2025-Semantic), (iii) a smoke test that runs end-to-end in $\sim$10 minutes on a single consumer GPU, and (iv) a reduced reproduction recipe for the inductive-only t-SNE ablation in \cref{tab:inductive_ablation} that runs in $\sim$2.5 hours on a single H100. Larger raw outputs (per-sample VLM responses for the main result table) are hosted on Hugging Face; the code repository is hosted on GitHub.

\subsection*{Artifact check-list (meta-information)}
\begin{itemize}\itemsep0pt\parsep0pt
\item \textbf{Algorithm:} MARVIS --- multi-modal classification/regression via VLM reasoning over t-SNE visualizations of pretrained embeddings.
\item \textbf{Program:} Python 3.11+ package \texttt{marvis}.
\item \textbf{Compilation:} \texttt{pip install -e ".[vision,audio,api]"}; no native compilation required.
\item \textbf{Models:} DINOv2-ViT-L/14-reg (vision), Whisper-large-v2 / Microsoft CLAP (audio), BioCLIPv2 (biological), TabPFNv2 (tabular), Qwen2.5-VL-3B-Instruct (VLM reasoning). All retrieved automatically from Hugging Face / torch.hub on first run.
\item \textbf{Data sets:} CC18-Semantic and Regression2025-Semantic (released with this artifact); CIFAR-10/100 (torchvision); ESC-50, RAVDESS, UrbanSound8K (audio benchmarks); fishnet, awa2, plantdoc (biological benchmarks).
\item \textbf{Run-time environment:} Linux x86\_64 or aarch64. Tested on Ubuntu 22.04 with CUDA 12.x and on macOS 14+ (Apple Silicon; transformers backend, MPS acceleration). VLLM optional and used automatically when available.
\item \textbf{Hardware:} For the smoke test, any consumer NVIDIA GPU with $\geq$8 GB VRAM (or Apple Silicon with $\geq$16 GB unified memory; transformers backend). For the inductive-ablation reproduction, one NVIDIA H100 (or comparable) is recommended. For the full benchmark suite (Table 1 in the paper), $\sim$1{,}500 H100-hours total.
\item \textbf{Metrics:} Classification accuracy (top-1), regression $R^2$, end-to-end and per-sample latency.
\item \textbf{Output:} Per-task accuracy/$R^2$ summaries (\texttt{*\_test\_results.json}, \texttt{*\_test\_summary.csv}), per-sample VLM prompts and responses, optional saved scatter-plot visualizations.
\item \textbf{Experiments:} Driven by shell-callable scripts in \texttt{examples/} and \texttt{marvis\_scripts/}; a Slurm \texttt{sbatch} script for the inductive ablation is included.
\item \textbf{How much disk space required:} 50 GB for the smoke test (includes pretrained model weights). 100--150 GB for the inductive ablation reproduction (adds CIFAR-10 image cache and per-sample plots/responses). $\sim$500 GB if mirroring all hosted result tarballs.
\item \textbf{How much time is needed to prepare workflow?} $\sim$30 minutes (clone, install, download model weights on first run).
\item \textbf{How much time is needed to complete experiments?} $\sim$10 minutes for the smoke test, $\sim$2.5 hours for the inductive-only t-SNE ablation reproduction (1 H100), days--weeks for the full benchmark sweep.
\item \textbf{Publicly available?} Yes. Code on GitHub; large result archives on Hugging Face Datasets; an archived release tag (DOI via Zenodo) is provided for the camera-ready commit.
\item \textbf{Code licenses:} MIT.
\item \textbf{Data licenses:} Released artifacts (CC18-Semantic, Regression2025-Semantic) are CC-BY 4.0. Third-party datasets retain their original licenses (e.g., CIFAR is MIT-style; OpenML CC18 follows OpenML terms).
\item \textbf{Workflow framework used?} No. Plain Python entry points and Slurm sbatch.
\item \textbf{Archived (provide DOI)?} GitHub release \texttt{v0.1.0-cais2026} archived on Zenodo: \texttt{<DOI to be filled in upon Zenodo deposit>}.
\end{itemize}

\subsection*{Description}

\subsubsection*{How to access}
\begin{itemize}\itemsep0pt\parsep0pt
\item \textbf{Code (small, mutable):} \url{https://github.com/penfever/marvis}, tag \texttt{v0.1.0-cais2026}.
\item \textbf{Code (archived, immutable):} Zenodo deposit, DOI \texttt{<TBD>}.
\item \textbf{Hosted datasets and result archives:} \url{https://huggingface.co/datasets/penfever/marvis} (CC18-Semantic, Regression2025-Semantic, and per-sample VLM outputs for the main results table).
\item \textbf{Reviewer entry point:} the README in the GitHub repository contains a section labeled \emph{``Artifact Reviewers --- Start Here''} that points to the smoke test and the inductive-ablation reproduction recipe.
\end{itemize}

\subsubsection*{Hardware dependencies}
A CUDA-capable GPU is recommended. The full benchmark sweep was conducted on H100 80GB nodes. The smoke test runs on any GPU with $\geq$8 GB VRAM and on Apple Silicon (Mac) under the transformers backend.

\subsubsection*{Software dependencies}
Python $\geq$3.11; PyTorch $\geq$2.0; \texttt{transformers}, \texttt{tabpfn}, \texttt{openml}, \texttt{torchvision}, \texttt{scikit-learn}, \texttt{matplotlib}. Full pinned requirements in \texttt{pyproject.toml}. Optional: \texttt{vllm} (Linux + CUDA only; falls back to \texttt{transformers} elsewhere).

\subsubsection*{Datasets}
CC18-Semantic and Regression2025-Semantic are introduced in this paper and released alongside the artifact (HF). All other datasets are downloaded automatically from their canonical sources on first run; no licensing exceptions are required.

\subsection*{Installation}
\begin{verbatim}
git clone https://github.com/penfever/marvis.git
cd marvis
conda create -n marvis python=3.11 -y
conda activate marvis
pip install -e ".[vision,audio,api]"
\end{verbatim}
On first invocation, MARVIS will fetch any missing pretrained embedding models (DINOv2, Whisper/CLAP, BioCLIPv2, TabPFNv2) and the Qwen2.5-VL-3B VLM into the local Hugging Face / \texttt{torch.hub} caches.

\subsection*{Experiment workflow}

\noindent\textbf{Smoke test ($\sim$10 minutes, single consumer GPU).} Verifies that the install is working end-to-end on a small subset of CIFAR-10:
\begin{verbatim}
python examples/vision/evaluate_all_vision.py \
    --datasets cifar10 \
    --models marvis_tsne \
    --balanced_few_shot \
    --num_few_shot_examples 10 \
    --max_test_samples 20 \
    --use_knn_connections --nn_k 5 \
    --zoom_factor 2.0 \
    --backend transformers \
    --vlm_model_id "Qwen/Qwen2.5-VL-3B-Instruct" \
    --output_dir ./smoke_results
\end{verbatim}
Expected runtime: $\sim$10 minutes after first-run model downloads. Expected output: \texttt{smoke\_results/cifar10/cifar10\_test\_results.json} containing accuracy on 20 test points; on a working install this is typically $\geq$60\%.

\noindent\textbf{Inductive-only t-SNE ablation reproduction ($\sim$2.5 hours, single H100).} Reproduces \cref{tab:inductive_ablation} (the main camera-ready ablation):
\begin{verbatim}
sbatch marvis_scripts/marvis-ablate-inductive-cifar.sbatch
\end{verbatim}
The sbatch sequentially runs a transductive-baseline condition followed by an inductive condition under matched settings (DINOv2-ViT-L/14-reg, $k=30$, zoom 15, Qwen2.5-VL-3B; full CIFAR-10 50k training set; 1{,}000 held-out test images). Expected runtime: $\sim$2.5 hours total. Expected outputs: per-condition \texttt{cifar10\_test\_results.json}.

\noindent\textbf{Full benchmark sweep.} The remaining sbatch scripts in \texttt{marvis\_scripts/} (\texttt{marvis-tabular-cls.sbatch}, \texttt{marvis-vision-marvis.sbatch}, \texttt{marvis-audio.sbatch}, \texttt{marvis-bio-marvis.sbatch}, plus the OpenML CC18 / Regression2025 orchestrators under \texttt{examples/tabular/openml\_*}) reproduce the rows of the main results table. The total compute budget is $\sim$1{,}500 H100-hours; per-sbatch budgets are documented in the README.

\subsection*{Evaluation and expected results}

\textbf{Smoke test.} On 20 CIFAR-10 test images with $k=10$ training samples per class and zoom factor 2.0, expected accuracy is in the 0.55--0.75 range (variance is dominated by the 20-sample test set). Receiving any non-error JSON output indicates a working install.

\textbf{Inductive-only t-SNE ablation.} On full CIFAR-10 training and 1{,}000 held-out test images, the camera-ready paper reports:
\begin{center}\small
\begin{tabular}{lcc}
\toprule
\textbf{Variant} & \textbf{Accuracy} & \textbf{t-SNE fit} \\
\midrule
Transductive (default) & 97.8\% & $\sim$210s \\
Inductive (LR projector) & 96.4\% & $\sim$170s \\
\bottomrule
\end{tabular}
\end{center}
A reproduction within $\pm$1 percentage point on each row is expected (variation comes from t-SNE initialization, ordering of test points within a batch, and stochastic decoding by the VLM).

\textbf{Full benchmark sweep.} Reproducing the full Table 1 of the main paper requires the compute budget noted above. We do not expect artifact reviewers to reproduce the full sweep; the inductive-ablation reproduction recipe above is the recommended scope for the \emph{Reproduced} badge.

\subsection*{Notes}

\begin{itemize}\itemsep0pt\parsep0pt
\item Compute nodes on shared HPC clusters (e.g., the Jupiter Booster used for the camera-ready ablation) often disable internet access. The provided sbatch script pre-stages CIFAR-10 and the DINOv2 and Qwen2.5-VL-3B weights via login-node Python helpers (commands in the README), and sets \texttt{HF\_HUB\_OFFLINE}, \texttt{TRANSFORMERS\_OFFLINE}, and \texttt{TORCH\_HOME} so the compute job uses local caches only.
\item On Apple Silicon, MARVIS auto-detects MPS and switches to the transformers backend; set \texttt{VLLM\_AVAILABLE=false} to force this if needed.
\item For the inductive variant, the linear-regression projector is intentionally the simplest plausible parametric approximation of t-SNE (closed-form least squares from training embeddings to training t-SNE coordinates). The accompanying paper section discusses the choice and frames the result as a lower bound on what an inductive deployment of MARVIS could achieve.
\item Per-sample VLM prompts and responses for the main result table (one JSON record per test point, $\sim$60 GB total) are mirrored to the Hugging Face dataset URL above and are not redistributed via GitHub.
\end{itemize}

\end{document}